%% file: manuscript.tex
\documentclass[manuscript,screen,nonacm]{acmart}

\usepackage{graphicx}
\usepackage{cleveref}
\usepackage{tabularx}
\usepackage{multirow}
\usepackage{fontawesome5} 
\usepackage{xcolor}
\usepackage{siunitx}

\AtBeginEnvironment{appendices}{\crefalias{section}{appendix}}

\setcopyright{none}
\copyrightyear{2025}
\acmYear{2025}
\acmDOI{XXXXXXX.XXXXXXX}
\acmConference[Conference acronym 'XX]{Make sure to enter the correct
  conference title from your rights confirmation email}{June 03--05,
  2018}{Woodstock, NY}

\begin{document}

\title{Analyzing Sustainability Messaging in Large-Scale Corporate Social Media}

\author{Ujjwal Sharma}
\email{u.sharma@uva.nl}
\orcid{0000-0003-0285-1303}
\affiliation{%
  \institution{University of Amsterdam}
  \city{Amsterdam}
  \country{The Netherlands}
}

\author{Stevan Rudinac}
\email{s.rudinac@uva.nl}
\orcid{0000-0003-1904-8736}
\affiliation{%
  \institution{University of Amsterdam}
  \city{Amsterdam}
  \country{The Netherlands}
}

\author{Ana Mićković}
\email{a.mickovic@uva.nl}
\orcid{0000-0003-4106-9111}
\affiliation{%
  \institution{University of Amsterdam}
  \city{Amsterdam}
  \country{The Netherlands}
}

\author{Willemijn van Dolen}
\email{W.M.vanDolen@uva.nl}
\orcid{0000-0001-7768-6539}
\affiliation{%
  \institution{University of Amsterdam}
  \city{Amsterdam}
  \country{The Netherlands}
}
\author{Marcel Worring}
\email{m.worring@uva.nl}
\orcid{0000-0003-4097-4136}
\affiliation{%
  \institution{University of Amsterdam}
  \city{Amsterdam}
  \country{The Netherlands}
}

\renewcommand{\shortauthors}{Sharma et al.}


\begin{abstract}
  In this work, we introduce a multimodal analysis pipeline that leverages large foundation models in vision and language to analyze corporate social media content, with a focus on sustainability-related communication. Addressing the challenges of evolving, multimodal, and often ambiguous corporate messaging on platforms such as \(\mathbb{X}\) (formerly Twitter), we employ an ensemble of large language models (LLMs) to annotate a large corpus of corporate tweets on their topical alignment with the 17 Sustainable Development Goals (SDGs). This approach avoids the need for costly, task-specific annotations and explores the potential of such models as ad-hoc annotators for social media data that can efficiently capture both explicit and implicit references to sustainability themes in a scalable manner. Complementing this textual analysis, we utilize vision-language models (VLMs), within a visual understanding framework that uses semantic clusters to  uncover patterns in visual sustainability communication. This integrated approach reveals sectoral differences in SDG engagement, temporal trends, and associations between corporate messaging, environmental, social, governance (ESG) risks, and consumer engagement.  Our methods---automatic label generation and semantic visual clustering---are broadly applicable to other domains and offer a flexible framework for large-scale social media analysis.
\end{abstract}




\maketitle

\section{Introduction}

In recent years, corporations have increasingly turned to social media platforms such as \(\mathbb{X}\), formerly Twitter, to disseminate strategic messages, engage with stakeholders, and project brand values. This shift has generated a rapidly expanding digital corpus of multimodal content---encompassing text, images, and videos---that offers valuable insights into organizational strategies and public perception. However, annotating and analyzing such content at scale presents significant challenges. The large volume of data, combined with abstract or highly stylized messaging and a lack of visual consistency across thematic content defined at high semantic levels (often associated with policy or promotional content), makes systematic analysis difficult. The inherently dynamic and unstructured nature of social media further complicates the identification of content aligned with specific attributes or indicators, particularly when themes are expressed through diverse and often inconsistent visual representations. For example, companies promoting concepts like ``safe and resilient societies'' may post images of cities, people, or explanatory graphics, yet these depictions rarely share clear visual patterns, making it difficult to define clear, a priori criteria for identifying and categorizing such content.

Compounding these issues is the evolving nature of social multimedia analysis tasks. Analyzing corporate sustainability communications, for instance, involves identifying content related to themes such as renewable energy, waste reduction, or frameworks like the UN Sustainable Development Goals. These tasks are not only labor-intensive, time-consuming, and context-dependent but are also complicated by the shifting nature of task-specific indicators, as corporate messaging evolves in response to emerging crises, market dynamics, and social movements. Consequently, fixed classification schemes risk becoming obsolete. Addressing these challenges requires systems capable of identifying high-level semantic concepts in large-scale social multimedia while remaining adaptable to new and evolving analysis tasks.

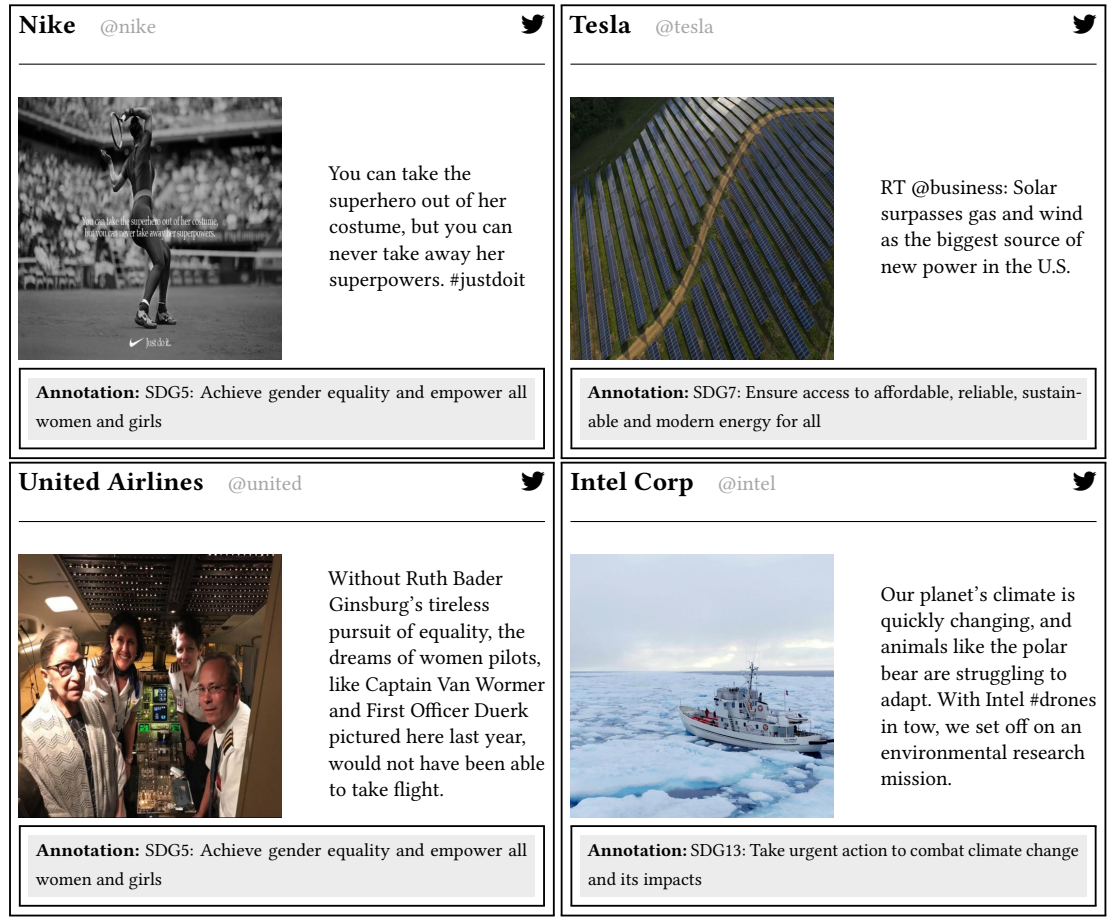
\begin{figure}
  \centering
  \input{example_tweets.tex}
  \Description{Corporate Tweets Annotated for Sustainability Relevance. The figure displays four examples of corporate tweets annotated to identify sustainability relevance. Each example includes an image, the tweet text, and an annotation indicating alignment with Sustainable Development Goals. 
        
    The first example is from Nike, showing a grayscale image of a woman in athletic wear. The tweet reads "You can take the superhero out of her costume, but you can never take away her superpowers. #justdoit". It is annotated as SDG5: Achieve gender equality and empower all women and girls. 
    
    The second example is from Tesla, featuring an image of solar panels. The tweet says ``RT @business: Solar surpasses gas and wind as the biggest source of new power in the U.S.''. This is annotated as SDG7: Ensure access to affordable, reliable, sustainable and modern energy for all. 
    
    The third example is from United Airlines, showing a color image of two women pilots. The tweet reads "Without Ruth Bader Ginsburg’s tireless pursuit of equality, the dreams of women pilots, like Captain Van Wormer and First Officer Duerk pictured here last year, would not have been able to take flight." It is annotated as SDG5: Achieve gender equality and empower all women and girls. 
    
    The final example is from Intel Corp, displaying a color image of drones flying over a snowy landscape. The tweet reads "Our planet’s climate is quickly changing, and animals like the polar bear are struggling to adapt. With Intel #drones in tow, we set off on an environmental research mission.” This is annotated as SDG13: Take urgent action to combat climate change and its impacts.
    }

  \caption{Examples of corporate tweets annotated using our methodology to identify sustainability relevance (annotations from our annotation approach are displayed below the tweet in a grey box). Our LLM-based approach reveals that explicit mentions of the Sustainable Development Goals (SDGs) are uncommon, with sustainability claims often conveyed through contextual information. This highlights a central component of this work: detecting sustainability-related content in corporate social media posts.}\label{fig:examples}
\end{figure}

Recent developments in large-scale foundation models offer compelling solutions to these challenges. These models are designed to be highly versatile, transferrable systems that serve as a base for numerous downstream applications. Language-centric models can generate coherent, contextually relevant text ranging from free text and code to nuanced responses in interactive dialogues. By leveraging vast pretraining, they excel at tasks such as translation, summarization, and question answering~\cite{brown2020language}. In parallel, foundation models for vision have similarly enabled a wide range of applications in computer vision~\cite{radford2021learning,ramesh2021zero,carion2020endtoend}. Notably, these models exhibit impressive zero-shot capabilities, enabling them to tackle new tasks through careful prompting rather than conventional retraining. Such models can be used to reason around complex themes, expressed visually or textually, in a zero-shot manner without extensive feature engineering.

In this work, we leverage the zero-shot properties of foundation models to design a multimedia analysis pipeline that combines large language models (LLMs) and vision-language models (VLMs) for comprehensive analysis of corporate social media content. Our approach addresses both textual and visual dimensions of communication, which are critical for shaping public perception and engagement. First, we use an ensemble of LLMs to annotate large-scale textual data (e.g., corporate tweets) by mapping their content to a thematic taxonomy and capturing both implicit and explicit references in the process. By aggregating results from multiple model families, we mitigate risks of task-specific performance variability and hallucinations, while enhancing accuracy and robustness in thematic tagging. Second, we extend this analysis to visual content---such as images and infographics---using representations derived from vision-language models. These models capture semantic and contextual relationships within images, enabling the identification of visually salient themes linked to key outcomes. By unifying text and visual analysis, our pipeline provides complementary perspectives of multimedia data, allowing practitioners to explore and interpret corporate social media content holistically. This dual approach not only ensures a robust, scalable evaluation of communication strategies but also establishes a foundation for targeted, cross-modal investigations into multimedia narratives and their impact.

We test these approaches to examine corporate communications on sustainability by aligning them with the United Nations Sustainable Development Goals (SDGs). By systematically categorizing tweets on their relevance to the SDGs, we gain insights into industry-level and company-specific engagement---such as which sectors most frequently discuss particular SDGs and how these discussions fluctuate over time. Moreover, the annotations allow deeper investigation into content-level patterns, revealing themes commonly used by specific industry sectors. Additionally, we use the aforementioned visual understanding pipeline to identify and analyze prevalent themes in visual content associated with these tweets. Some examples of the content examined in this work are provided in \Cref{fig:examples}. We summarize our contributions as follows:

\begin{figure}
  \centering
  \includegraphics[width=\textwidth]{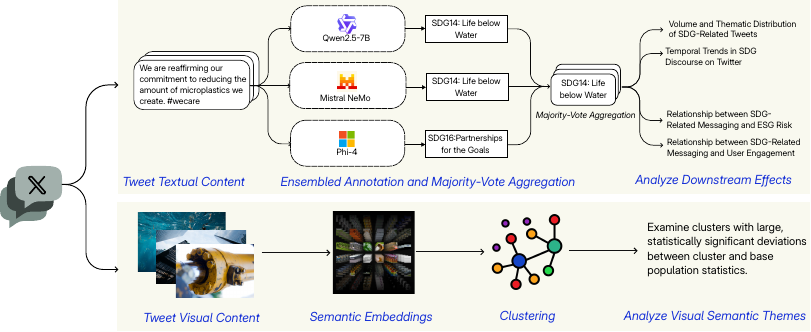}
  \Description{Proposed Approach for Analyzing Corporate Social Media Content

The figure illustrates a diagram outlining the proposed approach for analyzing corporate social media content related to sustainability. The process is divided into two main components: textual analysis and visual semantic theme analysis, both contributing to the overall objective of analyzing sustainability (SDG) aligned content in tweets. 

The first step is the textual analysis, starting with ``Tweet Textual Content''.  An example tweet reads ``We are reaffirming our commitment to reducing the amount of microplastics we create. #wecare.'' This content feeds into a multple large language models, specifically Qwen2.5 -7B, Mistral Nemo, and Phi-4. These models generate annotations which are then aggregated to assign Sustainable Development Goal classifications, specifically SDG14: Life below Water and SDG16: Partnerships for the Goals to the example tweet. The output of this stage is then used to analyze downstream effects, including volume and thematic distribution of SDG-related tweets, temporal trends in SDG discourse on Twitter, relationships between SDG-related messaging and Environmental, Social, and Governance risk, and relationships between SDG-related messaging and user engagement.

Parallely, the visual content of tweets,  represented by a collage of images, is processed to generate ``Semantic Embeddings'' which are then used for ``Clustering.'' This allows the analysis to ``Analyze Visual Semantic Themes,'' examining clusters with statistically significant deviations from base population statistics. 

The two analytical pathways converge to provide a comprehensive view of SDG-related tweet content across industries.}
  \caption{Overview of the proposed approach for analyzing corporate social media content. The pipeline includes two main components: (1) textual analysis using large language models to classify tweets by their relevance to the 17 Sustainable Development Goals (SDGs) and explore links to real-world outcomes such as ESG risk and engagement, and (2) visual analysis using a vision-language model to identify visual semantic themes tied to elevated ESG risk and/or engagement. Together, these analyses provide a comprehensive view of SDG-related tweet content across industries.}\label{fig:approach}
\end{figure}

\begin{enumerate}
  \item We propose an ensemble framework comprised of large language models (LLMs) serving as \textit{ad-hoc, zero-shot annotators} for analyzing social media text. This method leverages LLMs to robustly identify dimensions or attributes of interest (e.g., sustainability discourse) with minimal supervision, offering a scalable alternative to traditional manual annotation approaches.

  \item We introduce a novel bottom-up visual analytics framework to systematically assess deviations in performance metrics (e.g., sustainability outcomes, user engagement) between specific semantic themes and the broader population. This approach identifies visually salient concepts and themes that exhibit statistically significant deviations from baseline trends, enabling the discovery of hidden patterns in social media content.

  \item We operationalize these methods to analyze how major corporations communicate sustainability efforts, focusing on their alignment with the United Nations' Sustainable Development Goals (SDGs). By linking social media discourse to corporate sustainability indicators (e.g., Environmental, Social and Governance (ESG) risk scores) and audience engagement metrics, we uncover how firms' online narratives correlate with real-world sustainability performance and audience engagement.

  \item Applying our visual analytics framework, we examine semantic themes embedded in visual content (e.g., images, infographics) paired with corporate tweets across industries. We demonstrate how specific visual themes correlate with elevated ESG risk profiles or heightened user engagement, revealing the interplay between visual rhetoric and stakeholder perception in sustainability communication.
\end{enumerate}

The remainder of this work is organized as follows. \Cref{section:related_work} reviews relevant literature to contextualize our study. \Cref{section:preliminaries} introduces the Sustainable Development Goals (SDGs) and ESG risk as key preliminaries in our analysis. In \Cref{section:methodology}, we detail our two-stage methodology for analyzing textual and visual content. The first stage employs textual analysis 
of tweets, while the second stage leverages our proposed unsupervised visual understanding framework for identifying high-level themes associated with visual content of the tweets. \Cref{section:exp_setup} outlines the experimental design, including data sources, preprocessing steps, and evaluation metrics. \Cref{section:results} presents our findings and discusses their implications. Finally, \Cref{section:conclusion} summarizes the contributions and conclusions of this work.

\section{Related Work}\label{section:related_work}



Though multimedia predates social media, the emergence of large-scale social multimedia platforms such as $\mathbb{X}$ (formerly Twitter), Instagram, 
and others has catalyzed advancements in both domains. For example, social multimedia---multimodal content such as text, images, videos, or combinations thereof posted by users on social media platforms--- has been used to improve cross-modal representations for images and text~\cite{zhang2016learning}; discover 
images suitable for visual summaries using crowdsourced user preference signals~\cite{rudinac2013learning}; to discover the relationship between 
popularity bursts for video content and the social prominence of the video's underlying content on other social platforms~\cite{roy2013towards}; to 
examine the role of mined social and demographic indicators in travel recommendations and to infer the social contexts of users for publicity and 
marketing tasks~\cite{servia2013inferring}.  

Multimedia data from social platforms like $\mathbb{X}$ has emerged as a critical tool for social sensing, enabling researchers to track public health trends and the spread of illnesses like the flu~\cite{Paul_Dredze_2021}; to analyze how bots and social media content shape political discourse~\cite{Varol_Ferrara_Davis_Menczer_Flammini_2017}; to 
detect the spread of rumors and misinformation~\cite{dou2021user,he2021rumor,shang2023amica}; for fact-checking and bot detection~\cite{saeed2022crowdsourced,FengTwiBot20}; and for real-time threat detection in public infrastructure, such as metro systems~\cite{He2023}. 
While the aforementioned studies primarily focus on textual media, these analyses have been extended to visual media to analyze visual rhetoric in image and video advertisements~\cite{Hussain2017}; identify visual elements distinctive of specific 
geo-spatial areas~\cite{Doersch2012,10.1145/2531602.2531613}; validate predictive relationships between a city's visual attributes and non-visual 
outcomes~\cite{arietta2014city,Quercia_Schifanella_Aiello_McLean_2021}; determine visual components associated with livelier and safer 
neighborhoods~\cite{DeNadai2016,10.1145/2631775.2631799}; explore how high-level visual concepts are expressed in user-generated visual 
content~\cite{sharma2024pixels}; compare diffusion patterns across topics on social media~\cite{romero2011differences}; and delineate political 
communities and measure ideological divides~\cite{conover2011political}. As a dynamic, continuously updated repository of socially relevant information, social multimedia functions as a weakly annotated dataset offering 
insights into social dynamics and their wider societal implications.

The availability of social multimedia has not only facilitated social analyses but also driven advancements in multimedia research tools. Recent 
breakthroughs in visual representation learning (e.g., CLIP~\cite{radford2021learning} and BLIP~\cite{li2023blip}), image-generation models (e.g., 
DALL·E~\cite{ramesh2021zero}), and vision-language models have leveraged, among others, vast social media datasets to build foundational models that can generalize to a wide range of tasks.

With these advancements in hand, researchers are now exploring new application areas, such as sustainability-aligned communications. While existing 
research on sustainability has principally focused on the financial risk and reputational damage aspects~\cite{wong2022stock}, increasingly attention is 
shifting to \textit{content-centric} examinations of everyday corporate communications. For instance, 
\citet{supran2017assessing} explored patterns and indicators of greenwashing in curated social media posts from a limited set of companies, which were 
manually annotated to identify such markers. More recently, the \textit{GreenScreen} dataset~\cite{sharma2024greenscreen}, a key dataset used in this work, was introduced to systematically link corporate social media accounts with their associated tweets. This enables a broader analysis of corporate sustainability efforts by unifying social-media and contextual data. Despite growing momentum in research examining sustainability's societal, economic, and environmental implications, \emph{multimedia-centric studies} on this topic remain in their infancy and advancing the development of such techniques is a central objective of this work.

\section{Preliminaries}\label{section:preliminaries}

In this section, we discuss some key concepts essential for understanding the methods and analyses presented in this work. We begin with introducing the United Nations Sustainable Development Goals as a framework for sustainability, then introduce ESG risk as a proxy measure for sustainability. 

\subsection{Sustainable Development Goals (SDGs)}

Before examining sustainability themes in corporate social media content, it is critical to establish our framework for defining sustainability: the \emph{Sustainable Development Goals (SDGs)}. Adopted by the United Nations in 2015 as part of the 2030 Agenda for Sustainable Development~\cite{united-nations-2015}, these 17 interconnected objectives address urgent global challenges such as poverty, inequality, climate change, and environmental degradation. Simultaneously, they promote peace, justice, and equitable prosperity through shared targets and indicators, urging countries, organizations, and individuals to collaborate toward a sustainable future. A defining feature of the SDGs is their integration: progress in one area inherently influences others, reflecting the understanding that systemic challenges require holistic solutions.

Central to the SDGs' vision are three pillars of sustainability, as defined by the United Nations: \emph{environmental sustainability}, which focuses on safeguarding the planet's resources to ensure their availability for present and future generations; \emph{economic sustainability}, which emphasizes cultivating inclusive economic growth, decent work, and innovation while preserving resources and empowering communities; and \emph{social sustainability}, which prioritizes building equitable societies that uphold human rights, foster dignity, and guarantee security for all. These pillars are deeply interdependent. For instance, addressing climate change (environmental) may spur green job creation (economic) and improve public health (social). Conversely, neglecting social equity during environmental policymaking risks fueling inequality, destabilizing economies, and undermining long-term stability. Thus, while it may seem that environmental stewardship is the main focus of sustainability, the scope of the SDGs is far broader approach that harmonizes ecological, economic, and social priorities.

\subsection{ESG Risk as a Measure of Sustainability Performance}

The \textit{ESG (Environmental, Social, and Governance) risk score} is a composite indicator that reflects a company's potential exposure to various non-financial risks and its ability to manage those risks effectively. This score is obtained by combining both quantitative and qualitative assessments, taking into account factors such as the company's industry, geographical footprint, and other operational specifics.

A high ESG risk score indicates significant exposure to critical issues in one or more of the following areas: environmental risks, such as harm to natural resources, pollution, or vulnerability to stringent climate-related regulations; social risks, including problematic labor practices, strained community relations, or human rights challenges that could trigger public backlash or legal complications; and governance risks, such as weak oversight, poor management practices, or corruption-related issues that may undermine sound strategic decision-making. Overall, a high ESG risk score suggests that a company may be struggling to manage these key areas effectively, potentially leading to higher regulatory, reputational, and operational risks, thereby undermining its long-term sustainability. In this work, the ESG risk score serves as the principal measure for evaluating a company's sustainability-aligned risk.

\subsection{Supervisory Signals and Their Role in Our Work}

This work examines the properties and distribution of content posted by companies through a \emph{two-level hierarchical structure}. This structure consists of a finite set of \emph{groups} (e.g., companies) and a larger collection of \emph{constituent instances} (e.g., social media posts) associated with each parent. While this framework is generalizable---applicable to domains such as restaurants and their photos, or social media users and their posts---it is adapted here to analyze corporate social media content. In this hierarchy, information exists at both group and instance levels, with characteristics inherent to the group (e.g., corporate financial 
metrics, industry sector, or user metadata) and characteristics inherent to instances (e.g., engagement metrics like likes, retweets, 
or post content properties).

Specifically, in our work, we use \emph{ESG-risk scores}---which assess environmental, social, and governance risks associated with the companies---as the group-level information. At the instance level, we use engagement scores, derived from metrics like \emph{likes}, \emph{retweets}, or \emph{shares} of individual posts. These attributes are \emph{not used for model training} but instead serve as post-hoc analytical tools. Specifically, we use them to investigate whether observable properties or clusters of content (e.g., tone or thematic focus) at the instance level correlate with high or low values in either the ESG-risk or engagement metrics.

The choice of supervisory signals (e.g., ESG-risk or engagement metrics) is context-dependent and can be adapted to different analysis scenarios. For instance, in a financial analysis context, stock price volatility could replace ESG-risk scores, while in a user-centric study, follower growth might substitute engagement metrics. Critically, this flexibility does not compromise the core methodological framework of our work, which is agnostic to the specific signals chosen.


\section{Methodology}\label{section:methodology}

This study employs a two-pronged methodology to examine corporate communication on \(\mathbb{X}\). The first component analyzes textual content to map tweets to the 17 Sustainable Development Goals (SDGs), while the second component provides visual understanding of semantic themes encoded in visual content of the tweets. This dual approach allows us to capture both the narrative and the visual dimensions of corporate sustainability messaging. An overview of the proposed approach is provided in \Cref{fig:approach}.



\subsection{SDG Classification in Corporate Tweets via LLM Ensembles}\label{section:approach_part_one}

\subsubsection{Approach Rationale}
The textual analysis component of this work examines tweets to identify and categorize content according to the 17 Sustainable Development Goals (SDGs). By annotating tweets, we can systematically analyze tweet volumes, thematic patterns, and engagement metrics across companies and industry sectors. This approach provides a macroscopic view of corporate sustainability communication, enabling us to discern trends and shifts in how SDG-related content is prioritized and conveyed.

\subsubsection{Method} We begin by considering a collection of companies \(C = \{c_1, c_2, \ldots, c_m\}\), where each company \(c_i\) has its own set of tweets \(T_i = \{t_{i1}, t_{i2}, \ldots, t_{in_i}\}\). Each tweet includes textual content and engagement metrics (e.g., likes, retweets, and replies), and may optionally include media such as images or videos. Given the 17 Sustainable Development Goals (SDGs) denoted by \(SDGs = \{SDG_1, SDG_2, \ldots, SDG_{17}\}\), our objective is to design a mapping function
\begin{equation}
  f: T_i \to SDGs \cup \{\text{None}\},
\end{equation}
which assigns each tweet either to one of the 17 SDGs or to a \textit{None} category for tweets that do not pertain to any SDG---typically those focused on product promotion or financial disclosures.

A traditional approach might suggest training a supervised multi-class classifier. This classifier uses a set of tweets \emph{manually annotated} on relevance to the 17 SDGs to train a model that can then classify the remaining set of tweets. However, assembling a diverse dataset that fully captures SDG-relevant content is challenging due to the labor-intensive nature of annotation and the evolving, nuanced language of social media discourse. To overcome this limitation, we propose a strategy that leverages large language models (LLMs) to annotate tweets. Unlike traditional supervised models that require task-specific data annotations and training, LLM-based classifiers (like ours) draw on pre-existing knowledge obtained from vast web-scale data to generalize across novel tasks without explicit retraining. In particular, we use an LLM-based classifier that labels each tweet as one of 18 classes (17 SDGs plus the None category).

Since large language models (LLMs) may generate unreliable outputs (``hallucinations'') or exhibit inconsistent performance across diverse tweet types, we adopt a \emph{non-cascaded, post-inference ensembling approach} to mitigate these risks. This method involves simultaneously querying multiple LLMs from distinct model families with identical prompts and tweet inputs, then aggregating their predictions through majority voting~\cite{li2024more,si2023getting}. To ensure consistency and reduce noise from variability in the generated response (e.g., differing textual explanations for the same label), our prompts are designed to elicit a fixed set of responses (e.g., ``SDG3'' for SDG 3: Good Health) from each model. In cases of uncertainty or conflicting evidence, models are instructed to return a \textit{None} label. This uniform output format simplifies aggregation, enabling reliable majority voting with minimal post-processing.


We evaluate our LLM-based SDG-labeling framework using user-supplied hashtags as a ground-truth proxy. Hashtags have been widely adopted as annotations in Twitter studies (e.g., to analyze political discourse~\cite{conover2011political} or topic diffusion~\cite{romero2011differences}), and we leverage them here to infer SDG relevance. For each SDG category, we compile a set of core SDG hashtags (e.g., \texttt{\#SDG13} for SDG13) along with context-specific, thematically related hashtags (e.g., \texttt{\#CleanOceans}, \texttt{\#SavetheSeas}) to provide a comprehensive tagging system. The full list of these hashtags is provided in \Cref{appendix:sdg_hashtags}. Tweets containing these hashtags are processed by our framework to assess agreement between hashtag-derived ground truth and LLM-generated labels. Performance is evaluated individually for each LLM and collectively for the ensemble model (via majority voting), with inter-annotator consistency quantified using Cohen's Kappa and agreement percentages. 


The evaluation results guide our majority voting scheme. In the event of a tie---where multiple distinct judgments emerge---we use the computed performance scores to select the best-performing LLM as a tie-breaker. This strategy ensures that the final classification is not only consistent with the aggregated predictions but also benefits from the strengths of the most reliable model.

In summary, our approach integrates LLM-based annotations with an ensembling strategy and evaluates performance using proxy annotations from hashtags. This method aims to achieve a robust and scalable solution for mapping tweets to the Sustainable Development Goals while overcoming the limitations inherent in assembling large, annotated datasets.



\subsection{Extracting Collective Visual Themes in Corporate Social Media}\label{section:approach_part_two}

While the textual analysis reveals broad thematic trends, many companies also rely heavily on visual cues to reinforce their sustainability messages. The following section explores how these visual elements are organized and interpreted across different companies.

\subsubsection{Approach Rationale}

The textual analysis approach presented in~\Cref{section:approach_part_one} examines tweets to extract sustainability-related themes, but corporate communication on social media is not limited to text. Visual elements---such as images---play an equally important role in shaping public perception and engagement with sustainability topics. However, we deliberately adopt a different strategy for analyzing visual content.

Although visual content can be trivially linked to its corresponding tweet, engagement metrics, parent company, and ESG scores, a direct replication of the textual methodology would likely yield a parallel narrative without offering any additional insights. Instead, our goal is to uncover broad, overarching visual themes that emerge across the entire image collection. This approach allows us to identify key themes that underpin the sustainability messaging of high-risk companies or highly engaging tweets.

To achieve this, we considered two distinct analytical directions. A ``top-down'' approach would treat companies as the primary unit of analysis, focusing on the specific visual properties of their content. In contrast, a ``bottom-up'' approach first identifies visual themes \emph{within the complete image dataset} and then examines how individual companies contribute to or align with these common themes. In this work, we adopt the bottom-up approach, emphasizing the shared visual narratives that reveal how companies communicate sustainability beyond what is conveyed through text alone. A bottom-up approach is preferable for three main reasons:

\begin{enumerate}
  \item  Identifying visual themes that span multiple companies enables the discovery of patterns linked to specific risk profiles or high engagement. For instance, themes consistently seen in risky companies can inform the development of content-level indicators to supplement traditional risk analysis, while themes in highly engaging posts can provide actionable insights for corporate communication strategies.

  \item A top-down approach tends to isolate patterns within individual companies, which may not reoccur or hold broader relevance across industries. In contrast, a bottom-up method uncovers shared visual motifs and patterns that are applicable across different companies, offering more generalizable insights for experts and practitioners.

  \item When applying a top-down feature-attribution analysis, the risk of omitted-variable bias is high because company-level risk may cause models to conflate group-level signals with individual predictors. This can result in models that capture only company-specific characteristics rather than the broader, shared visual concepts. The bottom-up approach mitigates this issue by focusing on collective visual themes that provide richer and more interpretable results.
\end{enumerate}


\subsubsection{Method}

Let \(C = \{c_1, c_2, \ldots, c_m\}\) be the set of companies and let \(I = \{i_1, i_2, \ldots, i_n\}\) be the set of images (each corresponding to a tweet or post). Assume the function \(\phi : I \to C\) exists that maps each image \(i_j \in I\) to its associated company \(\phi(i_j) \in C\). At the company level, we obtain ESG-risk scores \(R(c)\) for each company \(c \in C\), which serve as a broad risk measure of the companies' sustainability performance. At the tweet level, we define an engagement measure \(E(i)\)---specifically the sum of like and retweet counts---for each image \(i \in I\), capturing the level of user engagement tied to the tweet.



Given these preliminaries, we seek a set of clusters \(K = \{K_1, K_2, \ldots, K_k\}\), where each cluster \(K_\ell\) is a subset of images. Given these clusters, we expect that clusters of interest should be (i) conceptually homogeneous (and therefore straightforward to interpret), (ii) widely shared (thus more likely to be generally applicable to a wide range of companies), and (iii) well-separated from the background population in terms of risk and/or engagement. Consequently, the properties captured by such clusters of interest may be consistently indicative of higher or lower risk or engagement profiles, thereby providing valuable insights for experts and practitioners. Formally, we operationalize the desiderata for clusters of interest as follows:

\begin{description}
\item [Semantic Coherence] Clusters of interest \(K_\ell\) should be internally coherent, containing a limited set of similar semantic concepts. Such coherence makes them easier to interpret and summarize. This can generally be ensured by selecting an appropriate clustering algorithm and tuning its hyperparameters effectively.

\item [High Diversity and Balanced Class Representation] Clusters of interest should span images from multiple companies (high diversity) while maintaining a balanced distribution of these companies (balanced representation), thereby highlighting concepts that are not idiosyncratic to one company but instead generalize across various companies. Formally, given the set of companies whose content appears in a cluster,
        \begin{equation}
          C(K_\ell) = \{\phi(i) : i \in K_\ell\},
        \end{equation}
        Our primary focus is on clusters exhibiting \emph{high diversity} (i.e., containing many distinct companies in $C(K_\ell)$) and \emph{balanced representation} of companies whose media appears in the cluster. This avoids scenarios where a large number of companies occupy a cluster but images from a single company predominantly monopolize the cluster. While the number of distinct companies in a cluster can be quantified via the cardinality of $C(K_\ell)$, its class balance can be assessed using a normalized form of Shannon's entropy~\cite{shannon1948mathematical}, defined as follows:
        \begin{equation}
          H_{\text{normalized}} = \frac{H}{\log(k)}
        \end{equation}

        Where \(H\) is Shannon's entropy and \(\log(k)\) is the maximum entropy, where \(k\) is the number of distinct companies within the cluster. In this case,\(H_{\text{normalized}} = 0\) exactly when the cluster is pure (all images belong to a single company). Conversely, \(H_{\text{normalized}} = 1\) when the cluster has a uniform class distribution (same number of images from each company). Together, the number of companies in a cluster and the normalized Shannon entropy provide an overall view of the size and distributional properties of the cluster.

\item [Large Separation from the Population] Clusters of interest are those in which constituent companies or images exhibit large, statistically significant deviations from the background population's medians with respect to their ESG-risk or engagement metrics. For each cluster \(K_\ell\), we define

        \begin{align}
          \widetilde{R}(K_\ell) & = \mathrm{median} \{ R(c) : c \in C(K_\ell) \}, \\
          \widetilde{E}(K_\ell) & = \mathrm{median} \{ E(i) : i \in K_\ell \}.
        \end{align}

        Let \(\widetilde{R}_{\text{pop}}\) and \(\widetilde{E}_{\text{pop}}\) be the \textit{medians} of risk and engagement for the background population. We then define the \emph{cluster risk deviation} and the \emph{cluster engagement deviation} as follows:

        \begin{align}
          \Delta_R(K_\ell) & = \widetilde{R}(K_\ell) - \widetilde{R}_{\text{pop}},  \\
          \Delta_E(K_\ell) & = \widetilde{E}(K_\ell) s- \widetilde{E}_{\text{pop}}.
        \end{align}

        We rank clusters by \(\Delta_R(K_\ell)\) or \(\Delta_E(K_\ell)\), which measure the extent to which the cluster's companies and images deviate from the background population. We deliberately do not use the absolute value because the sign of the deviation conveys crucial information. A large positive \(\Delta_R(K_\ell)\) indicates elevated risk (negative), whereas a large positive \(\Delta_E(K_\ell)\) indicates higher engagement (positive). Conversely, a large negative \(\Delta_R(K_\ell)\) implies reduced risk (positive), whereas a large negative \(\Delta_E(K_\ell)\) implies lower engagement (negative). While the magnitude of each deviation is relevant, its sign is essential to understanding the effect and interpretation of the deviation. To further confirm if the images are associated with companies that have a statistically different risk or if the images have a statistically different engagement, we run a Mann-Whitney U-Test~\cite{mann1947test} comparing the engagement or risk scores of each cluster's population to those of the background population. A cluster is only retained if the null hypothesis that the cluster's scores are the same as the background population's scores can be rejected with a p-value < 0.05. If this test does not pass, the cluster is not considered relevant for the particular score under evaluation.

\end{description}
\subsubsection{Discovering Visual Themes of Interest}
Given our formal structure and design criteria, our approach to discovering visual themes brings together these elements as follows. First, given a set of companies organized into sectors, we begin by selecting the specific sector to examine. Since our focus is on uncovering visual trends within a particular sector (for example, Materials, Energy, Finance, etc.), we designate the collection of images from all companies within the chosen sector as the background population for that task.

Next, the images associated with companies in this group are transformed into visual representation vectors using a vision-language model and then clustered. For each resulting cluster, we compute risk and engagement measures as well as the normalized Shannon entropy. We then rank the clusters based on their \(\Delta_R\) and \(\Delta_E\) values, and retain only those clusters where the statistical difference between the cluster and the overall population is confirmed using the Mann-Whitney U-Test~\cite{mann1947test}. This procedure yields clusters that are not only conceptually coherent and widely shared but also statistically distinct in terms of risk and engagement.

Finally, we analyze the visual content within these selected clusters to understand how companies communicate their sustainability efforts. This comprehensive analysis provides valuable insights into the visual language of sustainability and highlights how it varies across different companies and industries. The clusters resulting from this process often contain a large number of images, making manual examination both time-consuming and inefficient. To streamline this process for practitioners, we select a small subset of images from each cluster to serve as representative samples. Once the representative images are selected, they are processed by a generative vision-language model, which is prompted to generate two outputs: first, a single-line, high-level summary that encapsulates the overall theme of the images; and second, a list of key recurring themes or concepts, presented as short phrases. By repeating this process for every cluster, we obtain a collection of representative images along with their corresponding high-level summaries and key themes. This comprehensive approach not only enhances the interpretability of the clusters but also provides practitioners with a more user-friendly representation, facilitating a deeper understanding of the visual themes associated with different companies and industries.

\section{Experimental Setup}\label{section:exp_setup}

This section describes the data collection process and the experimental setup for the approaches presented in \Cref{section:approach_part_one} and 
\Cref{section:approach_part_two}.

\subsection{Data Collection}

We constructed our \(\mathbb{X}\) dataset using GreenScreen~\cite{sharma2024greenscreen}, a dataset providing Tweet IDs from a representative subset of Fortune 1000 companies. As GreenScreen only provides Tweet IDs, we utilized the \(\mathbb{X}\) API to rehydrate these, retrieving the full tweet content and associated metadata: tweet text, media, engagement statistics (retweets, replies, likes, and quote counts), and additional annotations. This resulted in a dataset of \emph{1,374,049 tweets} and \emph{715,081 accompanying images} posted between \emph{January 1, 2017} and \emph{December 13, 2022} by 537 distinct, active (i.e., non-zero posts) corporate Twitter accounts  and available ESG data.

In addition to tweet content, we incorporated fundamental company data---name, industry type, and stock symbol---obtained from GreenScreen. We then used these stock symbols to retrieve corresponding ESG risk scores from Sustainalytics, an ESG data provider~\cite{sustainalytics2024}. Finally, we classified each company into one of 10 \emph{sectors} based on the Global Industry Classification Standard (GICS)~\cite{gics2024}. These sectors are: \emph{Consumer Discretionary, Consumer Staples, Energy, Financials, Health Care, Industrials, Information Technology, Materials, Communication Services}, and \emph{Utilities}. Formal descriptions for these sectors are provided in \Cref{appendix:gics_sectors}.


\subsection{SDG Classification in Corporate Tweets via LLM Ensembles}\label{section:exp_setup_approach_part_one}

In \Cref{section:approach_part_one}, we outlined our approach for aligning tweets with the 17 Sustainable Development Goals (SDGs) using an LLM-based classifier. The proposed approach uses multiple LLMs to classify tweets based on their content, assigning each tweet either one of the 17 SDGs or \textit{None} category if irrelevant to the SDGs altogether.

Inspired by the chain-of-thought prompting paradigm introduced in \citet{wei2022chain}, we crafted a prompt that first provided a brief task overview, then detailed a step-by-step procedure for identifying the correct SDG. Rather than relying on specific examples (as in few-shot prompting), our focus was on the task's inherent structure, ensuring the model analyzed the tweet text itself. Given that the system's only role is as a classifier, we set this prompt as the system prompt before inference. The complete prompt is available in \Cref{appendix:sdg_classification_prompt}.

For annotation, we used \textit{Qwen2.5} (14B parameters, 5-bit quantization)~\cite{qwen2.5}, \textit{Mistral NeMo} (12B parameters, 6-bit quantization)~\cite{mistralnemo}, and \textit{Phi-4} (14B parameters, 4-bit quantization)~\cite{abdin2024phi}. These models offer a desirable tradeoff between accuracy, throughput (minimizing inference time on a massive tweet set) and memory constraints (via the inherent model size and varying levels of model quantization). We parsed each generated annotation, accepting it only if no extraneous reasoning or disclaimers were present; otherwise, we conservatively set the result to \textit{None}. We then aggregated these ensemble judgments.

For evaluation, we collected 6,310 tweets featuring hashtags that unambiguously map to the 17 SDGs. Based on the annotations agreement and Cohen's Kappa scores between the LLM-annotators and the label specified by the hashtags, the \textit{Qwen2.5} model was chosen as the tie-breaking annotator.

\subsection{Extracting Collective Visual Themes in Corporate Social Media}\label{section:exp_setup_approach_part_two}

In \Cref{section:approach_part_two}, we introduce an approach for clustering semantic vectors of visual content to uncover interesting properties, and this section details key implementation steps. First, we represent visual content using vector embeddings generated from a vision-language model trained with Sigmoid Loss for Language-Image Pre-Training on the WebLI dataset, as provided by the OpenCLIP project~\cite{zhai2023sigmoid,chen2022pali,ilharco_gabriel_2021_5143773}. To mitigate potential clustering bias, we remove duplicate images using perceptual hashing. Specifically, we employ the pHash algorithm~\cite{klinger2021phash} with a Hamming distance threshold of \textit{5} to identify and eliminate redundant content. Finally, images deemed irrelevant to the SDGs are filtered out, ensuring that the clustering analysis focuses solely on SDG-relevant visual content. Clustering is then performed on this cleaned dataset of images.

Given the large number of images in our dataset, selecting an appropriate clustering algorithm is critical. Techniques like K-Means require pre-specifying the number of clusters, a parameter that is unknown in our case and whose specification can dramatically alter the semantics of the discovered clusters. Although methods such as DBSCAN avoid this requirement, they struggle with density variations and are typically built on Euclidean distance, while our similarity measure is based on cosine distance.

To identify similar images, we implement a graph-based clustering algorithm on image embeddings. The process begins by normalizing embeddings and computing pairwise cosine similarities using batched matrix operations. A cluster forms around any point that has at least a minimum number of neighbors exceeding a predefined similarity threshold. These candidate clusters are extracted by identifying all neighbors above the threshold for qualifying points. The algorithm then sorts clusters by size and resolves overlaps by ensuring each point belongs to only one cluster, prioritizing assignment to larger clusters. In our specific operationalization of this clustering approach, we use a minimum cluster size of \emph{50} and a minimum cosine similarity threshold of \emph{0.75}, empirically chosen to ensure that the obtained clusters are semantically meaningful.

After clustering, representative images are sampled to capture key cluster concepts. We avoid random sampling as it risks imbalance in clusters with low normalized entropy and few distinct companies, where it may overrepresent a single company, misleading summarization by misidentifying a company's repeated presence as the cluster's primary theme. Instead, we use a round-robin strategy that first identifies all companies in a cluster, then iteratively samples one image randomly from each company until the required sample size is reached. This provides a more balanced representation, accurately reflecting the cluster's diversity.

The representative samples are then passed to a generative vision-language model---specifically, \emph{Qwen2.5-VL with 7 billion parameters}~\cite{Qwen2.5-VL}---along with a prompt instructing it to generate a high-level summary and list the predominant concepts and themes in the images. The exact prompt used for this task is provided in \Cref{appendix:visual_summary_prompt}.

\section{Results and Discussion}\label{section:results}

In this section, we present the results of our analysis based on the two-stage approach outlined in \Cref{section:approach_part_one} and \Cref{section:approach_part_two}. We begin by discussing the results obtained from the textual content analysis. We then use these annotations to discover and highlight trends in the tweets produced by companies, their ESG alignments, and the engagement metrics associated with these tweets. Finally, we present the results of the visual content analysis, focusing on the semantic clusters discovered and their implications for understanding corporate communication strategies.

\subsection{Evaluating the LLM-based Classifier}

To evaluate the LLM-based classifiers, we rely on the annotations they produce for a set of tweets associated with known hashtags, as described in \Cref{section:exp_setup_approach_part_one}. The results of this evaluation are presented in \Cref{tab:annotator_metrics}.

\begin{table}[!ht]
  \centering
    \caption{Agreement and Cohen's Kappa between LLM-generated SDG annotations and ground-truth annotations from associated hashtags. Majority Vote Aggregation uses Qwen2.5's judgement in case of ties.}
    \label{tab:annotator_metrics}
    \begin{tabular}{lll}
      \toprule
      Annotator                          & Agreement (\%)    & Cohen's Kappa    \\
      \midrule
      Qwen2.5                            & 80.21             & 0.77             \\
      Phi-4                              & 73.80             & 0.70             \\
      Mistral Nemo                       & 79.81             & 0.76             \\
      Majority Vote Aggregation & \underline{82.09} & \underline{0.79} \\
      \bottomrule
    \end{tabular}
\end{table}

From these results, we observe that among the individual LLM-based annotators, Qwen2.5 achieves the highest level of agreement and Cohen's Kappa with the hashtag-inferred annotations. Thus, we designate Qwen2.5 as the tie-breaking annotator in the majority-vote ensemble that determines the final label. The ensembled annotations exhibit the strongest performance across both metrics. It is also worth noting that the effectiveness of this ensemble approach depends strongly on the tie-breaking model: when a weaker model plays that role, the ensemble's agreement and Cohen's Kappa scores degrade noticeably. In the subsequent analyses, we use the annotations generated by majority-vote aggregation to study various aspects of corporate communications.

\subsection{Distribution of Overall and SDG-Relevant Tweet Volumes in Sectors}

\begin{figure}[htb]
  \centering
  \includegraphics[scale=0.5]{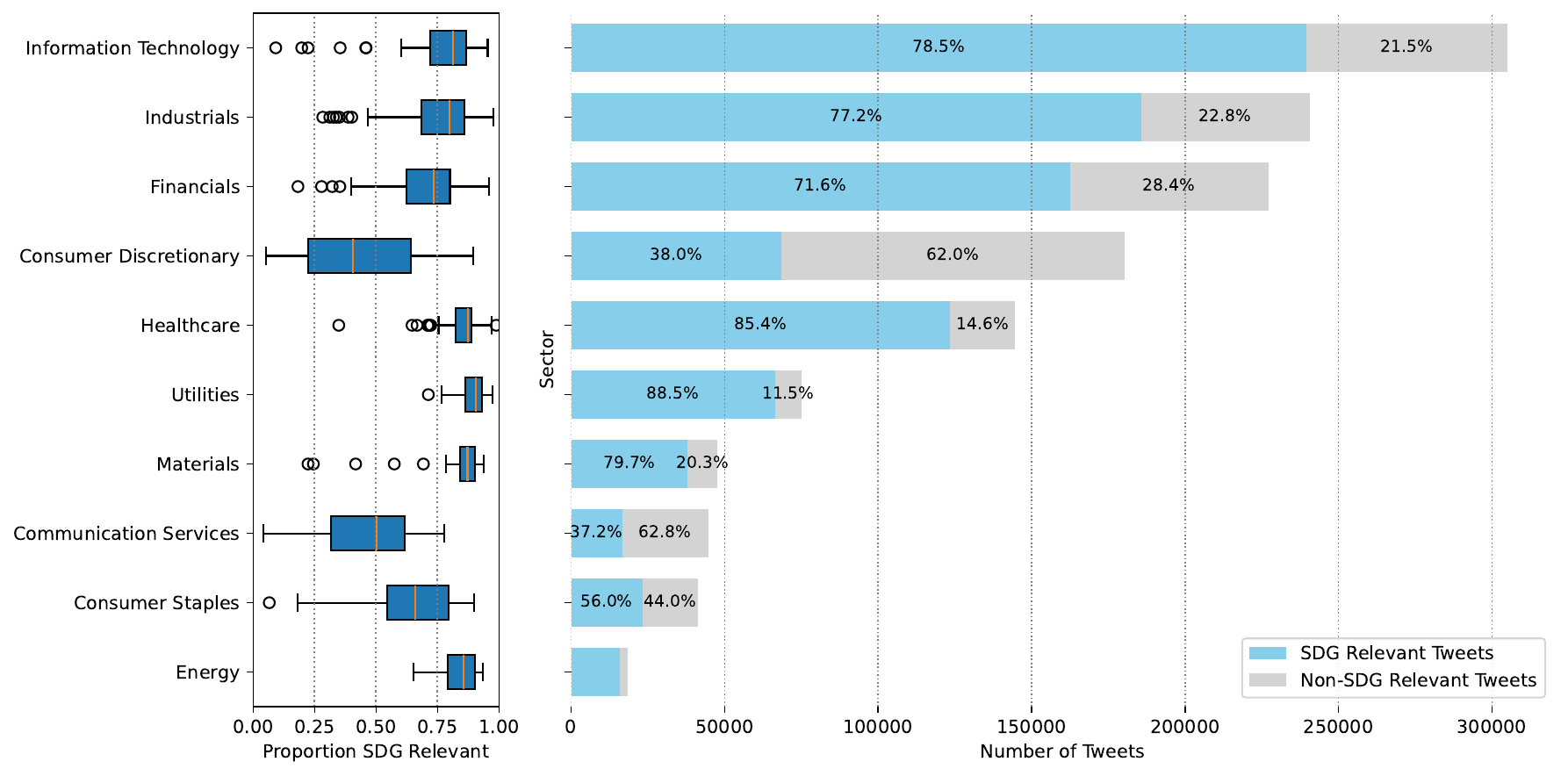}
  \Description{Total and SDG-relevant Tweet Volumes Across Various Sectors

The figure presents data on the proportion and volume of Sustainable Development Goal (SDG)-relevant tweets across different industry sectors. The left side displays boxplots illustrating the distribution of company-level proportions of SDG-relevant content within each sector, while the right side shows bar plots representing the aggregated proportion of SDG-relevant versus general content for each industry group.

The bar plots on the right provide a visual representation of the proportion of SDG-relevant tweets within each sector. Information Technology has approximately 78.5\% SDG-relevant content out of total tweets. Industrials show a SDG-relevant percentage of 77.2\%. Financials are around 71.6\%, Consumer Discretionary is 38.0\%, Healthcare shows 85.4\%, Utilities have 88.5\%, Materials display 79.7\%, Communication Services exhibit 37.2\%, Consumer Staples show 56.0\% and Energy has a high-value of SDG-relevant content, though the exact number is not disclosed due to visual space constraints. An interpretation of this image's conclusions is provided in the text.}
  \caption{Total and SDG-relevant Tweet Volumes Across Various Sectors. The boxplots on the left depict the distribution of company-level proportions of SDG-relevant to total content, providing a macro-averaged view of SDG-related communication within each industry. The bar plot on the right shows the aggregated proportion of SDG-relevant versus general content for each industry group.}
  \label{fig:tweetcounts-vs-industry}
\end{figure}

We first examine the proportion of tweets labeled as relevant to at least one SDG using the approach described in \Cref{section:approach_part_one}. Tweets with an SDG annotation (as computed by the majority vote) were classified as SDG-relevant; all others were deemed irrelevant. A summary of these proportions is provided in \Cref{fig:tweetcounts-vs-industry}. Our analysis indicates that the prevalence of SDG-related content varies considerably across different industries. In most sectors---excluding Communication Services, Consumer Discretionary, and Consumer Staples---the proportion of SDG-relevant content is substantially high, suggesting that SDG-aligned messaging has become woven into standard outward communication in these industries.

It is important to underscore that we measure content \emph{allied} with SDG themes, not explicit references to the SDGs themselves. This distinction is crucial because companies often embed sustainability-oriented information within business-relevant contexts rather than overtly referencing sustainability or the SDGs. We observe that industries traditionally facing greater sustainability challenges (e.g., Energy, Materials, and Utilities) exhibit a relatively high-mean and low-variance in the proportion of SDG-relevant content, indicating that consistently high levels of such communication are the norm. Meanwhile, sectors like Financials, Industrials, and Healthcare also show high mean values for SDG-related communication, albeit with more variability than in Energy, Materials, and Utilities.

\begin{figure}[htb]
  \centering
  \includegraphics[width=\linewidth]{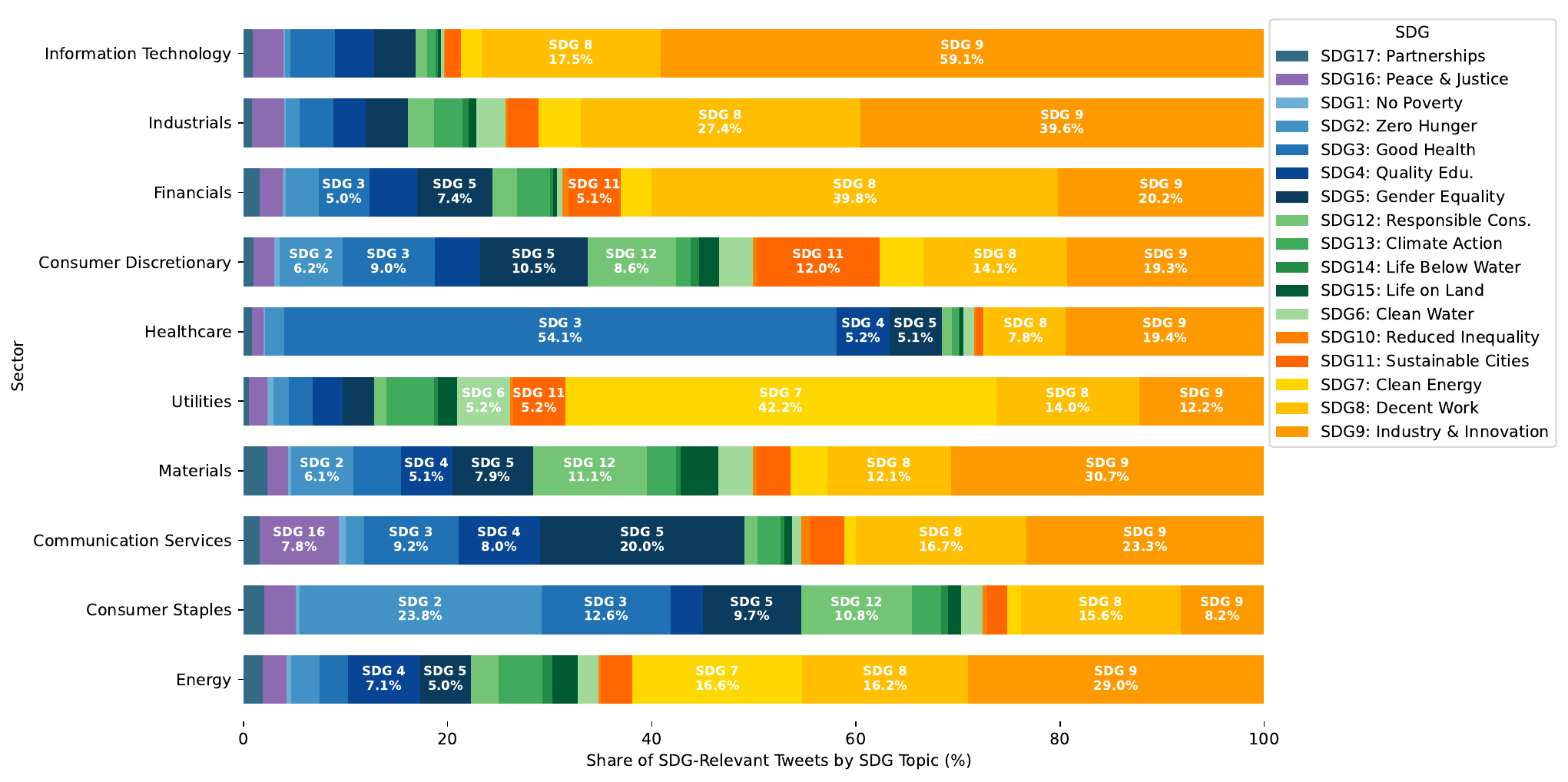}
  \Description{Sector Contributions to Sustainable Development Goals

This figure presents a stacked bar chart showing the share of each sector's tweets relevant to different United Nations Sustainable Development Goals, ranging from zero percent to one hundred percent. The sectors are listed vertically: Information Technology, Industrials, Financials, Consumer Discretionary, Healthcare, Utilities, Materials, Communication Services, Consumer Staples, and Energy. 

The horizontal axis represents the percentage share of each sector’s communications relevant to a specific Sustainable Development Goal. Each bar is divided into colored segments representing different goals, with a legend provided on the right side of the figure identifying each color with its corresponding goal number and name: SDG 1 – No Poverty, SDG 2 – Zero Hunger, SDG 3 – Good Health and Well-being, SDG 4 – Quality Education, SDG 5 – Gender Equality, SDG 6 – Clean Water and Sanitation, SDG 7 – Affordable and Clean Energy, SDG 8 – Decent Work and Economic Growth, SDG 9 – Industry, Innovation and Infrastructure, SDG 10 – Reduced Inequalities, SDG 11 – Sustainable Cities and Communities, SDG 12 – Responsible Consumption and Production, SDG 13 – Climate Action, SDG 14 – Life Below Water, SDG 15 – Life on Land, SDG 16 – Peace, Justice and Strong Institutions, and SDG 17 – Partnerships for the Goals. An interpretation of this image's conclusions is provided in the text.}

  \caption{Normalized distribution of SDG-related tweet occurrences, categorized by color-coded parent themes: `People' (SDGs 1--5; shades of blue), `Planet' (SDGs 6, 12--15; shades of green), `Prosperity' (SDGs 7--11; shades of yellow and orange), `Peace' (SDG 16; purple), and `Partnership' (SDG 17; teal). Each color group reflects a distinct overarching dimension of sustainable development.}
  \label{fig:normalized-sdg-distribution}
\end{figure}

Our analysis of the normalized distribution of the share of SDG-relevant tweets in \Cref{fig:normalized-sdg-distribution} reveals that sectors tailor their communication strategies to align with both operational priorities and stakeholder expectations. Notably, Information Technology, Financials, and Industrials concentrate heavily on SDG 8 (Decent Work and Economic Growth) and SDG 9 (Industry, Innovation and Infrastructure), with the IT sector exhibiting an extreme bias toward innovation. In contrast, while Energy, Materials, and Utilities also emphasize these economic themes, Utilities uniquely allocate a significant share to SDG 7 (Clean Energy), indicating a strategic commitment to sustainability. Healthcare predominantly focuses on SDG 3 (Good Health and Well-being), yet it also allocates a modest portion of its messaging to broader social dimensions such as education (SDG 4) and gender equality (SDG 5). Similarly, Communication Services and Consumer Staples extend their narrative beyond economic imperatives by incorporating SDG 5 (Gender Equality) and SDG 2 (Zero Hunger), respectively. An intriguing pattern emerges among sectors facing considerable environmental risks: despite their vulnerability, these industries tend to underrepresent direct environmental protection measures (e.g., climate action), instead emphasizing the positive socio-economic impacts of their operations. Overall, this distribution suggests a deliberate strategic choice by sectors to highlight positive work and innovation outcomes, while gaps in environmental discourse may indicate areas for potential future engagement.


\begin{figure}[htb]
  \centering
  \includegraphics[width=0.8\textwidth,keepaspectratio]{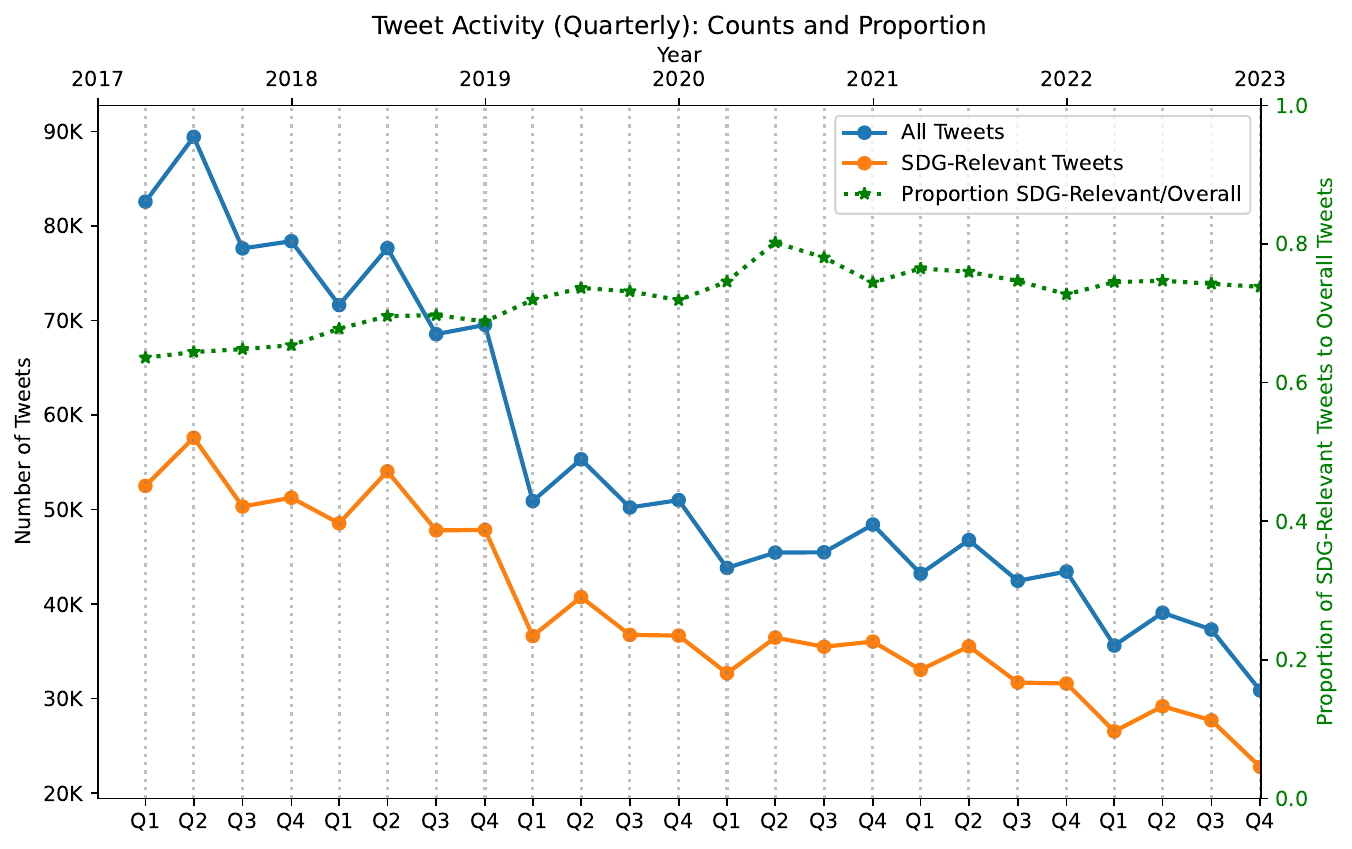}
  \Description{Tweet Activity Trends: Counts and Proportion

This line graph illustrates trends in tweet activity from January 2017 to December 2022, presented quarterly. The vertical axis on the left shows the number of tweets, ranging from 20,000 to 90,000. The vertical axis on the right displays the proportion of Sustainable Development Goal relevant tweets relative to overall tweets, ranging from zero to one.

Three lines are plotted: a blue line representing all tweets, an orange line depicting the absolute volume of Sustainable Development Goal relevant tweets, and a green dotted line showing the proportion of Sustainable Development Goal relevant tweets compared to total tweets. The horizontal axis indicates the quarter (Q1-Q4) for each year from 2017 to 2022. Generally, all tweet volumes decreased over time, with fluctuations observed in each quarter. The proportion of Sustainable Development Goal relevant tweets rose slightly throughout the period, fluctuating between approximately 0.2 and 0.6.}
  \caption{The lines indicate the evolution of total tweet volumes and SDG-relevant tweet volumes from January 2017 to December 2022. Solid lines depict absolute volumes of total and SDG-relevant tweets, while the dotted line shows the proportion of SDG-relevant tweets relative to total tweets. The quarter indicated by the x-axis labels shows the end of the quarter for which the data belonged.}
  \label{fig:overall-sdg-relevant-tweet-activity}
\end{figure}

\Cref{fig:overall-sdg-relevant-tweet-activity} depicts how overall tweet volumes and SDG-relevant tweet volumes evolved over time. While SDG-relevant posts generally track the broader trends in total tweets, the proportion of SDG-relevant tweets has steadily increased since 2017. This rising share suggests that companies are progressively integrating sustainability themes into their communications, reflecting a growing recognition of communicating on these topics.

\begin{figure}[htb]
  \centering
  \includegraphics[width=0.8\linewidth]{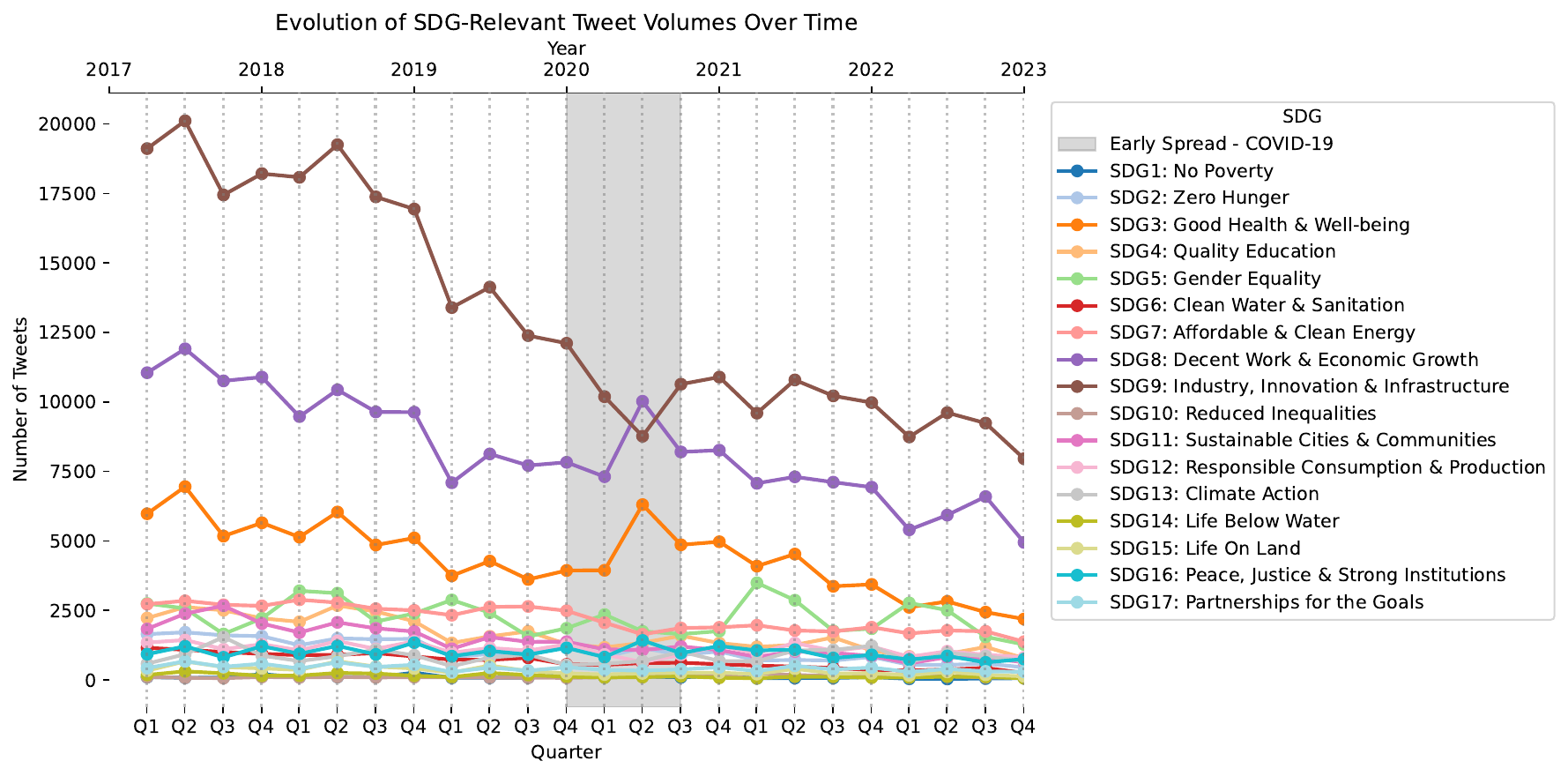}
  \Description{Evolution of Sustainable Development Goal-Related Tweet Volumes Over Time

The figure shows a line graph illustrating the absolute volumes of tweets related to each Sustainable Development Goal over time, from the first quarter of 2017 through the fourth quarter of 2022. The y-axis represents the number of tweets, ranging from zero to twenty thousand. The x-axis indicates the quarter, labeled as q1 through q4.

Each line on the graph corresponds to a specific Sustainable Development Goal, with different colored lines representing each goal. A shaded area highlights the time period corresponding to the early spread of COVID-19. 

The graph demonstrates fluctuations in tweet volumes for each SDG over the observed timeframe. For example, tweets related to `Decent Work and Economic Growth' (SDG8) and `Good Health and Well-being' (SDG3) show significant peaks around the second quarter of 2020, coinciding with the start of the COVID-19 pandemic. Other SDGs exhibit varying trends, some showing gradual increases or decreases in tweet volume over time.}
  \caption{The lines represent the absolute volumes of tweets for each SDG over time, showing the evolution of various SDG themes within the SDG-relevant volumes over time. The shaded area indicates the time-period corresponding to the early spread of COVID-19. The quarter indicated by the x-axis labels corresponds to the end of the quarter for which the data belonged.}
  \label{fig:sdg-relevant-tweet-activity}
\end{figure}

\Cref{fig:sdg-relevant-tweet-activity} offers a more granular perspective on the evolution of SDG-specific themes within the SDG-relevant tweet volumes. We observe that SDG 8 (Decent Work and Economic Growth) and SDG 9 (Industry, Innovation and Infrastructure) consistently appear among the most frequently mentioned topics, indicating a strong focus on themes related to economic and industrial growth. Conversely, other SDGs, such as SDG 13 (Climate Action), have not experienced a similar uptick, suggesting that companies may be prioritizing certain SDGs more than others.

A notable pattern appears between the end of Q4 2019 (December 31, 2019) and the end of Q3 2020 (September 30, 2020), a period coinciding with the early spread of COVID-19 and associated lockdowns. During these months, communications regarding SDG 3 (Good Health and Well-being) and SDG 8 (Decent Work and Economic Growth) spiked, underscoring heightened concerns over public health and economic uncertainty. Other themes received comparatively less attention during this time, reflecting the immediate focus on health risks and financial challenges.

\subsection{Relationship between SDG-Relevant Content and ESG Risk}

As mentioned earlier, ESG (Environmental, Social, and Governance) risk scores serve as numerical indicators of a company’s exposure to—and management of—various sustainability issues. A high ESG risk score signals potential weaknesses in these critical areas, which may undermine an organization’s long-term resilience. To explore the relationship between SDG-focused messaging and ESG risk, we compute the Spearman correlation between the proportion of tweets addressing a specific SDG (relative to total tweet volume) and the corresponding ESG risk scores. Each correlation value indicates whether increased emphasis on a particular SDG theme is associated with elevated (or reduced) ESG risk. We present a condensed version of the significant correlations in \Cref{tab:sdg-industry}, with the complete correlation table provided in \Cref{appendix:correlations}.

\begin{table}[htb]
  \centering
  \small
  \caption{Statistically significant correlations \((p < 0.05)\) between sector-level tweet volume focused on Sustainable Development Goals (SDGs) and ESG risk scores indicating which SDGs have the strongest relationship with ESG risk across different industry sectors. For each SDG, its number is shown in square brackets and the correlation coefficient in parentheses.}
  \label{tab:sdg-industry}
  \begin{tabularx}{\linewidth}{l X}
    \toprule
    Sector               & Top 5 Significant SDGs                                                                                                                                                                      \\ \midrule
    Communication Services & None                                                                                                                                                                                        \\ \midrule
    Consumer Discretionary & None                                                                                                                                                                                        \\ \midrule
    Consumer Staples       & None                                                                                                                                                                                        \\ \midrule
    Energy                 & Climate Action[13] (0.5789), Gender Equality[5] (0.4825)                                                                                                                                           \\ \midrule
    Financials             & Industry, Innovation, and Infrastructure[9] (-0.3432), No Poverty[1] (0.3301), Decent Work and Economic Growth[8] (0.3037), Quality Education[4] (0.2960), Affordable and Clean Energy[7] (-0.2895)        \\ \midrule
    Healthcare             & Clean Water and Sanitation[6] (-0.4191), Responsible Consumption and Production[12] (-0.3895), Good Health and Well-being[3] (0.3579), Industry, Innovation, and Infrastructure[9] (-0.3128)             \\ \midrule
    Industrials            & Responsible Consumption and Production[12] (-0.2947), Industry, Innovation, and Infrastructure[9] (0.2747), Zero Hunger[2] (-0.2682), Decent Work and Economic Growth[8] (-0.2135), No Poverty[1] (-0.2045) \\ \midrule
    Information Technology & Climate Action[13] (-0.3190), Gender Equality[5] (-0.2846), Good Health and Well-being[3] (-0.2839), Life on Land[15] (-0.2823), Quality Education[4] (-0.2723)                                              \\ \midrule
    Materials              & Decent Work and Economic Growth[8] (0.3404), Responsible Consumption and Production[12] (-0.3336)                                                                                                  \\ \midrule
    Utilities              & Sustainable Cities and Communities[11] (0.4427)                                                                                                                                                 \\ \bottomrule
  \end{tabularx}
\end{table}

Across sectors, there is a clear divergence in how higher ESG risk correlates with SDG-related communication. Notably, the Energy industry exhibits strong positive correlations with Climate Action (0.5789) and Gender Equality (0.4825), suggesting that as ESG risk increases, companies within this sector emphasize both environmental and social dimensions in their messaging. In contrast, Communication Services, Consumer Discretionary, and Consumer Staples show no significant relationships, potentially reflecting either broad risk-mitigation strategies or lower sensitivity to specific SDG themes. Within Financials, negative correlations for Industry, Innovation, and Infrastructure (-0.3432) and Affordable and Clean Energy (-0.2895) point to reduced attention to these topics among higher-risk firms, while positive correlations for No Poverty (0.3301), Decent Work and Economic Growth (0.3037), and Quality Education (0.2960) indicate that firms with greater exposure are more vocal about these social and economic SDGs. 

Healthcare similarly demonstrates mixed trends, with negative correlations for Clean Water and Sanitation (-0.4191) and Responsible Consumption and Production (-0.3895), but strong positive correlations for Good Health and Well-Being (0.3579), suggesting that high-risk healthcare firms focus on core domain strengths (health services) while avoiding environmental goals that may be harder to implement or communicate credibly within the sector.

Industrials, meanwhile, show predominantly negative correlations---suggesting less communication on Zero Hunger (-0.2682), Decent Work (-0.2135), and No Poverty (-0.2045), possibly reflecting an ESG strategy focused on reducing visibility as risk increases, with companies choosing to communicate less about sustainability topics. The exception is Industry, Innovation, and Infrastructure (0.2747), which may be highlighted as part of broader efforts to modernize industrial operations. Overall, this pattern suggests a cautious or defensive approach to sustainability communication. In Materials, higher-risk companies communicate more frequently on Decent Work (0.3404) but less on Responsible Consumption (-0.3336). Lastly, Utilities demonstrates a substantial positive correlation for Sustainable Cities and Communities (0.4427), implying that utilities facing greater ESG challenges emphasize sustainable infrastructure solutions. Taken together, these patterns point to nuanced, sector-specific strategies whereby higher-risk firms either reinforce certain SDGs central to their business models or downplay others that may lie outside their core operational focus.

The observed correlations between ESG risk and SDG-related communication suggest that higher-risk companies often emphasize SDGs that are symbolically positive or easier to align with public narratives, such as Climate Action and Gender Equality in Energy, or Decent Work and Quality Education in Financials, while downplaying goals that are harder to address or less directly tied to their operations. This selective engagement may reflect a strategic use of sustainability messaging to manage perceptions and signal commitment, even when underlying practices may not align. The lack of significant correlations in sectors like Consumer Staples or Communication Services further highlights that some industries may engage less with SDG narratives altogether, either due to lower public pressure or less perceived reputational benefit.

\begin{table}[htb]
  \centering
    \caption{Engagement Statistics for SDG-Relevant and Non-SDG-Relevant Tweets. Outliers were removed using Tukey's Fences prior to calculating the mean and median.}
    \label{tab:engagement-stats}
    \begin{tabular}{@{}lcccc@{}}
      \toprule
      \multirow{2}{*}{} & \multicolumn{2}{c}{Likes} & \multicolumn{2}{c}{Retweets}                 \\
      Tweet Classification                  & Mean                      & Median                       & Mean & Median \\ \midrule
      SDG-Relevant      & 4.27                      & 3.00                         & 1.93 & 1.00   \\
      Non-SDG-Relevant  & 8.19                      & 3.00                         & 3.06 & 2.00   \\ \bottomrule
    \end{tabular}
\end{table}

We also investigated whether SDG-relevant tweets receive higher user engagement, measured by likes and retweets, than non-SDG-relevant posts. \Cref{tab:engagement-stats} presents the distributions of engagement metrics for both categories. To assess whether the differences in engagement are 
statistically significant, we used the Mann-Whitney U test. However, we did not find statistically significant evidence to support greater engagement 
for SDG-relevant posts, for either likes or retweets.  

\subsection{Semantic Clusters of Visual Content}

\input{plates/materials_plate.tex}
\input{plates/financials_plate.tex}

In \Cref{section:approach_part_two}, we described our method for clustering semantic vectors of visual content to reveal salient visual attributes in visual content accompanying tweets. Here, we present the results of that analysis, emphasizing the clusters discovered and their implications for corporate communication strategies. As an illustrative example, we present findings for two sectors: \textit{Materials} and \textit{Financials}. The Materials sector, encompassing sub-industries such as oil and gas production and mining, faces broad sustainability risks due to its potential for environmental harm. In contrast, the Financials sector faces a mixed bag of risks from environmental, social and governance (ESG) issues. We selected these sectors to illustrate the diversity of visual themes and communication strategies across industries.

To identify and select clusters for further examination, we used the \(\Delta_R\) (risk deviation) and \(\Delta_E\) (engagement deviation) metrics, which capture how much each cluster’s median risk and engagement differ from those of the overall population. We retained only those clusters whose risk or engagement deviations were statistically significant according to a Mann-Whitney U-Test. Among these, we chose the top-two clusters based on \(\Delta_R\) and \(\Delta_E\), ensuring they contained at least five distinct companies and had a minimum normalized entropy of 0.3.

\subsection{Visual Themes in the Materials Sector}

The Materials industry clusters are shown in \Cref{fig:materials_plate}. We observe that clusters with a substantially higher risk deviation---reflected by large positive \(\Delta_R\)---tend to feature ``green'' themes such as tree planting and gardening. In the first cluster, such environmental themes are combined with community-oriented imagery, whereas the second cluster uses open fields and farming scenes to convey similar ideas. Although these clusters both exhibit a large risk deviation (+11.62 for the first cluster), we find that community-focused themes (first cluster) also enjoy higher engagement than the more agrarian imagery (second cluster).

The final cluster shows that tweets featuring imposing shots of oil and gas infrastructure can also achieve a notable positive engagement deviation (+9.00). This suggests that more responsible companies in this domain appear comfortable showcasing their core facilities, while higher-risk firms more frequently highlight community-centric content.

In the Materials sector, an interesting pattern emerges where companies with higher ESG risk are more frequently associated with visuals showing nature-related activities and community involvement, such as gardening, tree planting, or volunteer work. While these themes signal care for the environment and local communities, they are often only loosely related to the companies' core indutrial activities. In contrast, lower-risk firms tend to post more directly about their operations, including images of facilities, infrastructure, or production sites. This suggests that higher-risk companies might be using softer, symbolic imagery to shift attention away from their environmental and governance challenges. The contrast highlights a potential gap between how companies present themselves and the underlying realities of their business models, especially when ESG risk is high.

\subsection{Visual Themes in the Financials Sector}
In \Cref{fig:financials_plate}, we see that higher-risk companies in the Financials sector also emphasize community-centric themes, such as food distribution and donations. These themes are associated with a higher risk deviation (+7.19) but also generate elevated engagement deviation. Similarly, stock market-oriented imagery is extremely well-received, with a strong engagement deviation (+18.00), although it corresponds to a moderately increased risk deviation (+5.66). Finally, pro-LGBTQ+ themes stand out for garnering remarkably high engagement deviation (+15.00), though they similarly present a positive risk deviation above the background population.

In the Financials sector, high-risk companies often emphasize socially resonant themes, such as community support, food distribution, and LGBTQ+ inclusion, in their visual messaging. These themes, while publicly appealing, are typically peripheral to the firms' core financial activities. In contrast, companies with lower or moderate ESG risk are more likely to post visuals related to the financial domain itself, such as stock market boards and trading imagery. This suggests that high-risk firms may be leaning on symbolic social themes to enhance public perception, whereas lower-risk firms appear more confident in communicating their operational identity. The resulting contrast points to a strategic use of visual content as reputational signaling, particularly among firms facing higher ESG scrutiny. 

It is important to note that these outcomes, while semantically meaningful, have received minimal semantic filtering. The fact that themes such as ``green'' imagery and socially conscious content arise naturally---without being pre-specified---underscores the effectiveness of our bottom-up visual understanding approach. By discovering clusters prior to applying any thematic labels, we avoid the limitations of methods that begin by selecting over- or under-performing themes and then searching for related visual content.

\section{Conclusion}\label{section:conclusion}

In this work, we have presented a two-stage approach leveraging an ensemble of large-language models as \textit{ad hoc} annotators for large-scale social media analysis. This approach opens new possibilities for examining themes and motifs within such collections. We operationalize this procedure within the context of corporate communication aligned with the 17 United Nations Sustainable Development Goals, demonstrating how different sectors communicate about these goals and how their emphasis varies based on their overall sustainability risks---allowing us to identify thematic focus across different industry sectors. To further discover content-centric properties, we also propose a visual understanding approach that extracts thematic clusters from visual content and examines whether these clusters exhibit statistically meaningful deviations from the background population. Clusters with large deviations reveal interesting trends that could highlight potentially relevant behaviors in corporate communication strategies. Overall, these results demonstrate the strength of our techniques as a content analysis tool. In the context of our sustainability focus, our findings indicate that companies are increasingly integrating sustainability themes into their communications, with a significant variation in emphasis on specific SDGs across industries and distinct sector-specific themes evident in visual content.


\bibliographystyle{ACM-Reference-Format}
\bibliography{references}

\newpage
\appendix
\input{appendices.tex}

\end{document}

%% file: example_tweets.tex
\newcommand{\tweetcardFancy}[6]{
    \setlength{\fboxsep}{3pt}
    \setlength{\fboxrule}{0.7pt}
    
    \fbox{%
        \begin{minipage}{\dimexpr\linewidth-2\fboxsep-2\fboxrule\relax}
            \textbf{\large #2} \hspace{0.5em} {\small\textcolor{gray!70}{#3}} \hfill {\textcolor{black}\faTwitter} \\[1pt]
            \rule{\linewidth}{0.4pt}\\[0.25pt]
            
            \begin{minipage}[]{0.5\linewidth}
                \includegraphics[width=\linewidth, keepaspectratio]{#1}
            \end{minipage}%
            \hfill
            \begin{minipage}[]{0.41\linewidth}
                \small\raggedright #4
            \end{minipage}
            
            \vspace{2pt}
            \fbox{%
                \begin{minipage}{\dimexpr\linewidth-2\fboxsep-2\fboxrule\relax}
                    \colorbox{gray!15}{%
                        \begin{minipage}{\dimexpr\linewidth-2\fboxsep\relax}
                            {\footnotesize\textcolor{black}{\textbf{Annotation:} #6}}
                        \end{minipage}%
                    }%
                \end{minipage}%
            }%
        \end{minipage}%
    }%
}

\begin{center}
    \begin{minipage}{0.48\textwidth}
        \tweetcardFancy{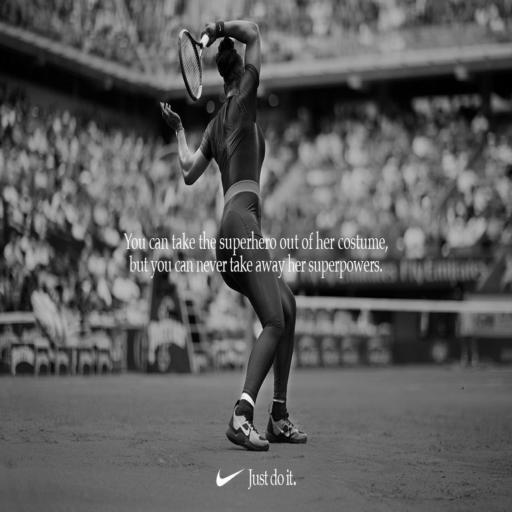}{Nike}{@nike}{You can take the superhero out of her costume, but you can never take away her superpowers. \#justdoit}{4:34 PM · Aug 25, 2018}{SDG5: Achieve gender equality and empower all women and girls}
    \end{minipage} 
    \begin{minipage}{0.48\textwidth}
        \tweetcardFancy{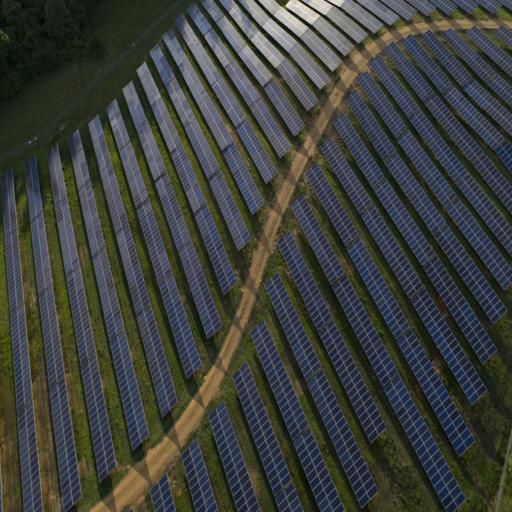}{Tesla}{@tesla}{RT @business: Solar surpasses gas and wind as the biggest source of new power in the U.S.}{8:17 PM · Jun 13, 2018}{SDG7: Ensure access to affordable, reliable, sustainable and modern energy for all}
    \end{minipage}
    \hfill
    \begin{minipage}{0.48\textwidth}
        \tweetcardFancy{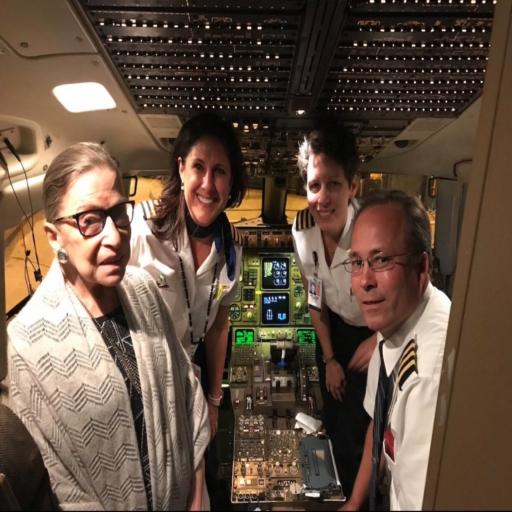}{United Airlines}{@united}{Without Ruth Bader Ginsburg's tireless pursuit of equality, the dreams of women pilots, like Captain Van Wormer and First Officer Duerk pictured here last year, would not have been able to take flight.}{10:13 PM · Sep 23, 2020}{SDG5: Achieve gender equality and empower all women and girls}
    \end{minipage} 
    \begin{minipage}{0.48\textwidth}
        \tweetcardFancy{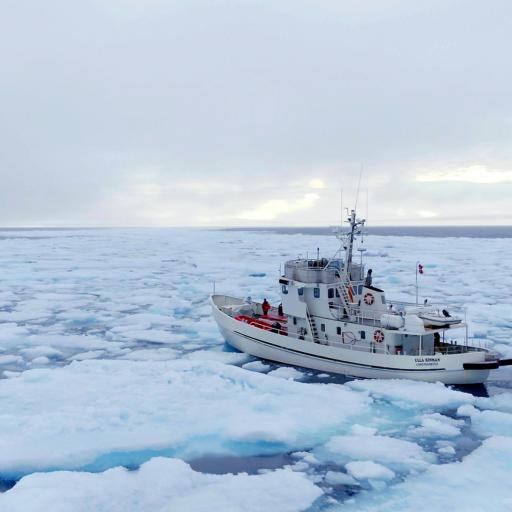}{Intel Corp}{@intel}{Our planet's climate is quickly changing, and animals like the polar bear are struggling to adapt. With Intel \#drones in tow, we set off on an environmental research mission.}{11:55 AM · Nov 16, 2020}{SDG13: Take urgent action to combat climate change and its impacts}
    \end{minipage}
    \hfill
\end{center}

%% file: plates/materials_plate.tex
\begin{figure}[htbp]
    \centering
    %
    \begin{minipage}[]{0.6\linewidth}
        \centering
        \includegraphics[width=\linewidth]{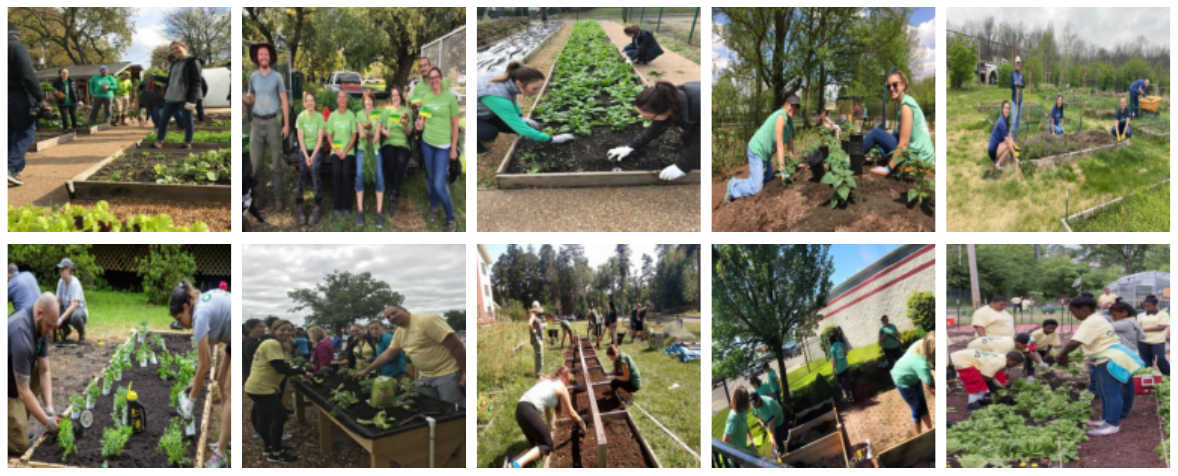}
    \end{minipage}
    \begin{minipage}[]{0.35\linewidth}
        \centering
        \small
        \begin{tabularx}{\linewidth}{l >{\raggedright\arraybackslash}X}
            \toprule
            \textbf{Property}                    & \textbf{Value}  \\ \midrule
            $\Delta_R$ (Delta Median Risk)       & \textbf{+11.62} \\
            $\Delta_E$ (Delta Median Engagement) & \textbf{+2.00}  \\
            Distinct Companies in Cluster        & 15              \\
            Normalized Cluster Entropy           & 0.64            \\
            \bottomrule
        \end{tabularx}
    \end{minipage}

    \vspace{1em}
    \parbox{0.9\linewidth}{
        \small\textbf{Associated Visual Concepts:} \\
        Community gardening, Environmental conservation, Volunteerism, Planting trees, Gardening activities, Environmental education, Community engagement, Sustainable practices, Outdoor work, Teamwork
    }

    \vspace{1em}
    \hrule
    \vspace{1em}

    %
    \begin{minipage}[]{0.6\linewidth}
        \centering
        \includegraphics[width=\linewidth]{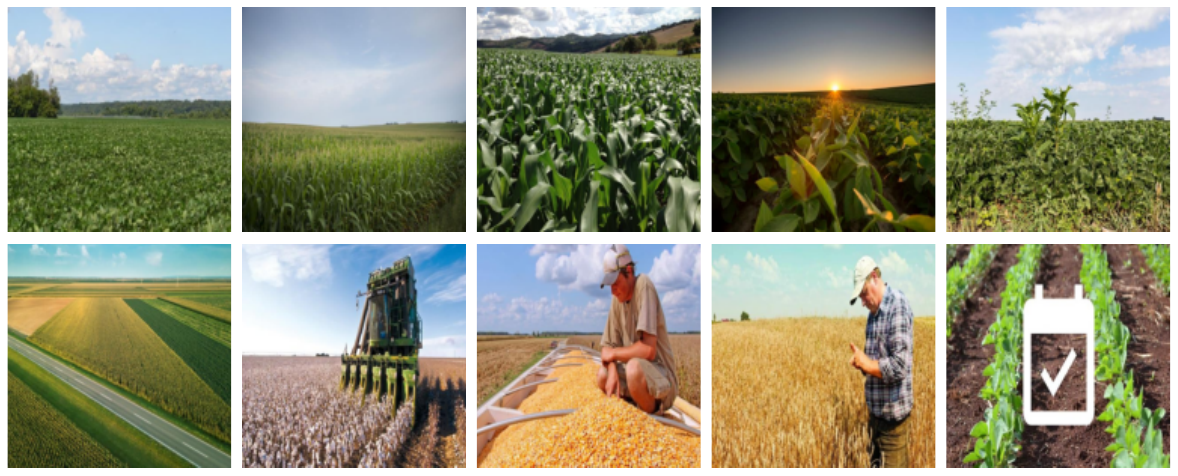} 
    \end{minipage}
    \begin{minipage}[]{0.35\linewidth}
        \centering
        \small
        \begin{tabularx}{\linewidth}{l >{\raggedright\arraybackslash}X}
            \toprule
            \textbf{Property}                    & \textbf{Value} \\ \midrule
            $\Delta_R$ (Delta Median Risk)       & \textbf{+8.28} \\
            $\Delta_E$ (Delta Median Engagement) & +0.00          \\
            Distinct Companies in Cluster        & 12             \\
            Normalized Cluster Entropy           & 0.70           \\
            \bottomrule
        \end{tabularx}
    \end{minipage}

    \vspace{1em}
    \parbox{0.9\linewidth}{
        \small\textbf{Associated Visual Concepts:} \\
        Farming activities, Harvesting, Crop cultivation, Agricultural machinery, Field work, Soil preparation, Planting, Crop inspection, Agricultural landscapes, Sustainable farming
    }




    \vspace{1em}
    \hrule
    \vspace{1em}

    %
    \begin{minipage}[]{0.6\linewidth}
        \centering
        \includegraphics[width=\linewidth]{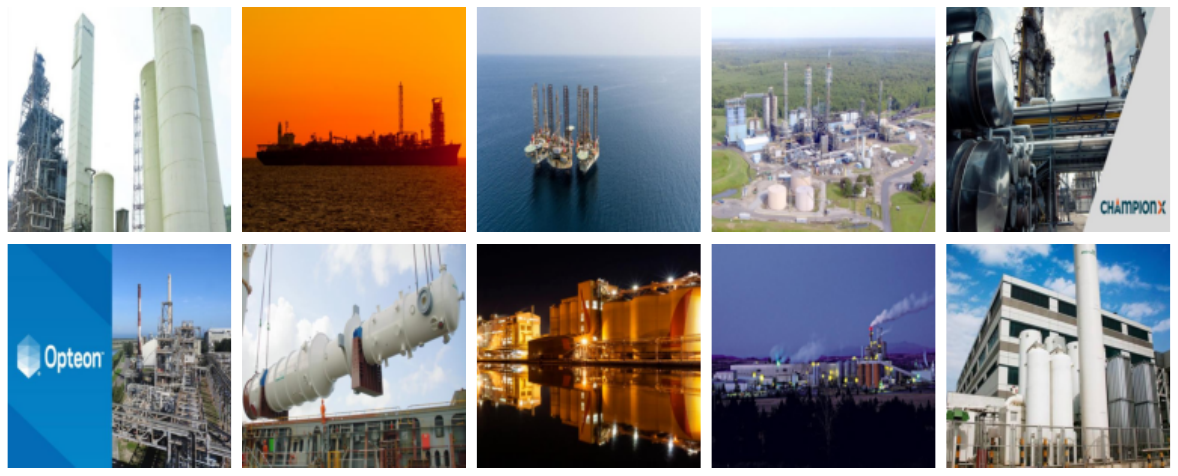}
    \end{minipage}
    \begin{minipage}[]{0.35\linewidth}
        \centering
        \small
        \begin{tabularx}{\linewidth}{l >{\raggedright\arraybackslash}X}
            \toprule
            \textbf{Property}                    & \textbf{Value} \\ \midrule
            $\Delta_R$ (Delta Median Risk)       & \textbf{-2.15} \\
            $\Delta_E$ (Delta Median Engagement) & \textbf{+9.00} \\
            Distinct Companies in Cluster        & 15             \\
            Normalized Cluster Entropy           & 0.83           \\
            \bottomrule
        \end{tabularx}
    \end{minipage}
    
    \vspace{1em}
    \parbox{0.9\linewidth}{
        \small\textbf{Associated Visual Concepts:} \\
        Oil refineries, Petrochemical plants, Power generation facilities, Pipeline networks, Manufacturing sites, Energy production, Industrial landscapes, Infrastructure development, Environmental impact
    }
    \Description{Visual Analysis of Three Clusters in the Financials Industry

The figure presents three distinct visual clusters related to the financials industry, each accompanied by associated visual concepts and statistical properties. The top cluster depicts a scene involving food distribution, clothing donation, painting, and renovation activities, suggesting community support or disaster relief efforts. Associated properties include a delta median risk of +7.19, a delta median engagement of +2.00, 20 distinct companies within the cluster, and a normalized cluster entropy of 0.86.

The middle cluster shows a parade scene with numerous participants holding flags and banners. The associated visual concepts relate to financial institutions, stock exchanges, small business celebrations, corporate events, and corporate branding. The statistical properties are a delta median risk of +5.66, a delta median engagement of +18.00, 20 distinct companies within the cluster, and a normalized cluster entropy of 0.79.

The bottom cluster illustrates a display of rainbow flags and imagery associated with LGBTQ+ pride, community support, equality, love, and celebration. The associated properties are a delta median risk of +3.32, a delta median engagement of +15.00, 23 distinct companies within the cluster, and a normalized cluster entropy of 0.91.
}
    \caption{Qualitative and statistical analysis of clusters in the \textbf{Materials} industry. Highlighted clusters represent the top three with the highest $\Delta_R$ and $\Delta_E$ values. Values in bold indicate statistically significant differences determined by a one-sided Mann-Whitney U test at the 0.05 significance level.}
    \label{fig:materials_plate}
\end{figure}

%% file: plates/financials_plate.tex
\begin{figure}[htbp]
    \centering

    %
    \begin{minipage}[]{0.6\linewidth}
        \centering
        \includegraphics[width=\linewidth]{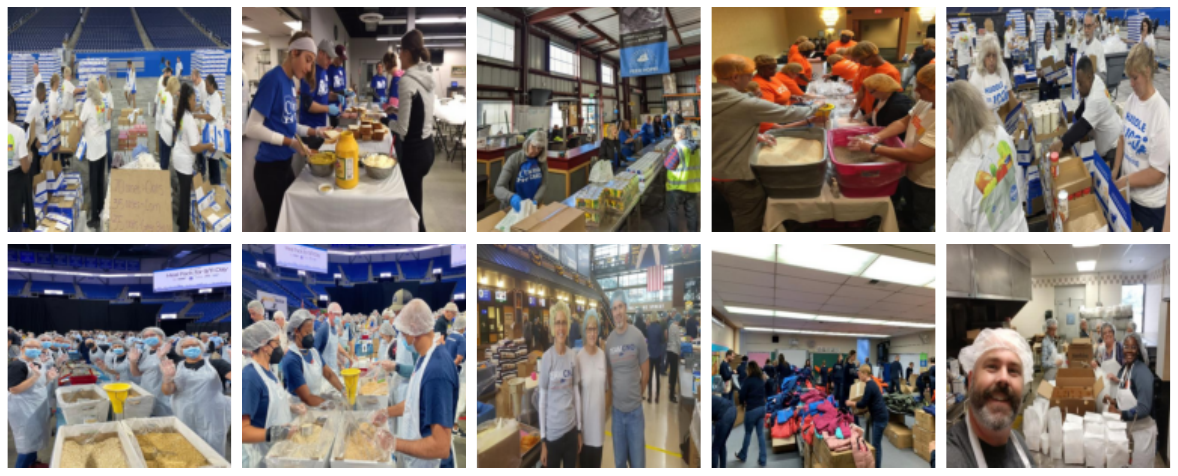}
    \end{minipage}
    \begin{minipage}[]{0.35\linewidth}
        \centering
        \small
        \begin{tabularx}{\linewidth}{l >{\raggedright\arraybackslash}X}
            \toprule
            \textbf{Property}                     & \textbf{Value} \\ \midrule
            $\Delta_R$ (Delta Median Risk)       & \textbf{+7.19} \\
            $\Delta_E$ (Delta Median Engagement) & \textbf{+2.00} \\
            Distinct Companies in Cluster       & 20             \\
            Normalized Cluster Entropy           & 0.86           \\
            \bottomrule
        \end{tabularx}
    \end{minipage}

    \vspace{1em}
    \parbox{0.9\linewidth}{
        \small\textbf{Associated Visual Concepts:} \\
        Food distribution, Home construction, Clothing donation, Painting and renovation, Event organization, Educational materials preparation, Community garden, Fundraising event, Disaster relief
    }

    \vspace{1em}
    \hrule
    \vspace{1em}

    %
    \begin{minipage}[]{0.6\linewidth}
        \centering
        \includegraphics[width=\linewidth]{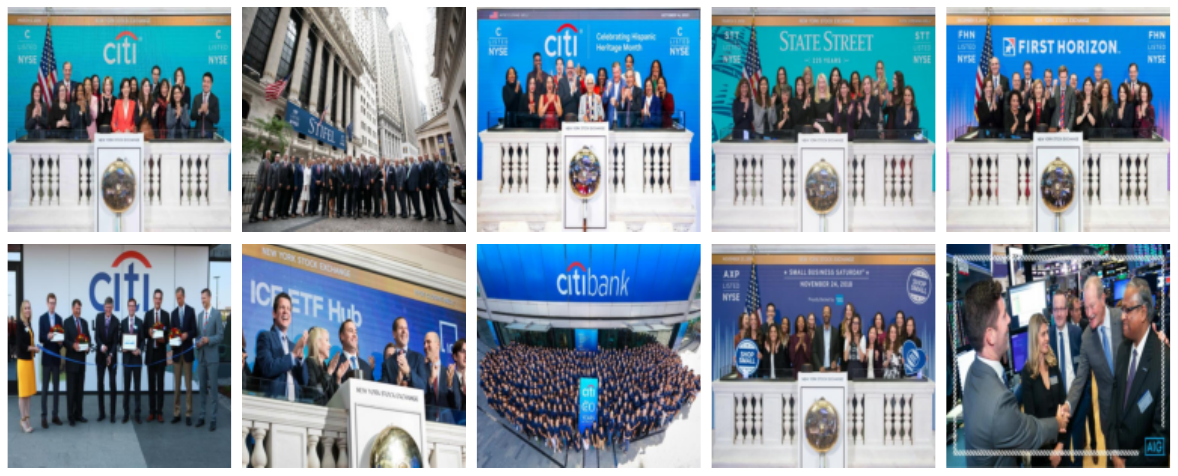} 
    \end{minipage}
    \begin{minipage}[]{0.35\linewidth}
        \centering
        \small
        \begin{tabularx}{\linewidth}{l >{\raggedright\arraybackslash}X}
            \toprule
            \textbf{Property}                     & \textbf{Value} \\ \midrule
            $\Delta_R$ (Delta Median Risk)       & \textbf{+5.66} \\
            $\Delta_E$ (Delta Median Engagement) & \textbf{+18.00} \\
            Distinct Companies in Cluster       & 20             \\
            Normalized Cluster Entropy           & 0.79           \\
            \bottomrule
        \end{tabularx}
    \end{minipage}

    \vspace{1em}
    \parbox{0.9\linewidth}{
        \small\textbf{Associated Visual Concepts:} \\
        Financial Institutions, Stock Exchanges, Small Business Celebrations, Corporate Events, Groundbreaking Ceremonies, Professional Gatherings, Community Involvement, Corporate Branding
    }


    %


    \vspace{1em}
    \hrule
    \vspace{1em}

    %
    \begin{minipage}[]{0.6\linewidth}
        \centering
        \includegraphics[width=\linewidth]{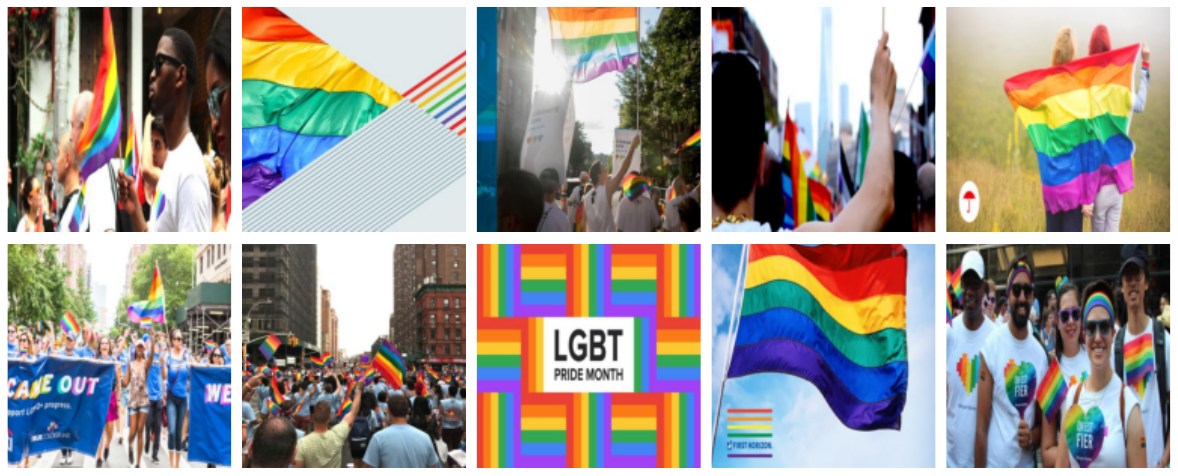}
    \end{minipage}
    \begin{minipage}[]{0.35\linewidth}
        \centering
        \small
        \begin{tabularx}{\linewidth}{l >{\raggedright\arraybackslash}X}
            \toprule
            \textbf{Property}                     & \textbf{Value}  \\ \midrule
            $\Delta_R$ (Delta Median Risk)       & \textbf{+3.32}  \\
            $\Delta_E$ (Delta Median Engagement) & \textbf{+15.00} \\
            Distinct Companies in Cluster       & 23              \\
            Normalized Cluster Entropy           & 0.91            \\
            \bottomrule
        \end{tabularx}
    \end{minipage}

    \vspace{1em}
    \parbox{0.9\linewidth}{
        \small\textbf{Associated Visual Concepts:} \\
        Pride Month, LGBTQ+ pride, Rainbow flags, Parade participants, Community support, Equality, Love, Celebration
    }
    \Description{Visual Analysis of Three Clusters in the Materials Industry

The figure presents three distinct visual clusters related to the materials industry, each accompanied by associated visual concepts and statistical properties. The top cluster depicts a group of people engaged in community gardening activities, including planting trees and working outdoors. Associated properties include a delta median risk of +11.62, a delta median engagement of +2.00, 15 distinct companies within the cluster, and a normalized cluster entropy of 0.64.

The middle cluster shows an agricultural scene with farming activities such as harvesting and crop cultivation, featuring agricultural machinery in a field. The associated properties are a delta median risk of +8.28, a delta median engagement of 0.00, 12 distinct companies within the cluster, and a normalized cluster entropy of 0.70.

The bottom cluster illustrates an industrial landscape with oil refineries, petrochemical plants, power generation facilities, and pipeline networks. The associated properties are a delta median risk of -2.15, a delta median engagement of +9.00, 15 distinct companies within the cluster, and a normalized cluster entropy of 0.83}
    \caption{Qualitative and statistical analysis of clusters in the \textbf{Financials} industry. Highlighted clusters represent the top three with the highest $\Delta_R$ and $\Delta_E$ values. Values in bold indicate statistically significant differences determined by a one-sided Mann-Whitney U test at the 0.05 significance level.}
    \label{fig:financials_plate}
\end{figure}

%% file: appendices.tex
\section{Hashtags corresponding to each SDG}\label{appendix:sdg_hashtags}

The hashtags listed in \Cref{tab:sdg_hashtags} correspond to each Sustainable Development Goal (SDG) and were used to assess the quality of the LLM-based SDG annotations.
\begin{table}[!htbp]
    \centering
    \small
    \caption{SDG Numbers, Their Titles, and Relevant Hashtags}
    \label{tab:sdg_hashtags}
\resizebox{0.9\textwidth}{!}{%
        \begin{tabular}{c l >{\raggedright\arraybackslash}p{0.45\textwidth}}
            \toprule
            \textbf{SDG Number} & \textbf{SDG Name}                       & \textbf{Relevant SDG Hashtags}                                                                             \\
            \midrule
            1                   & No Poverty                              & \texttt{SDG1, NoPoverty, EndPoverty, PovertyEradication, SocialProtection, FinancialInclusion, EconomicEmpowerment} \\
            \cmidrule{1-3}
            2                   & Zero Hunger                             & \texttt{SDG2, ZeroHunger, FoodSecurity, EndHunger, NutritionMatters, SustainableAgriculture, FoodForAll}            \\
            \cmidrule{1-3}
            3                   & Good Health and Well-Being              & \texttt{GoodHealth, HealthForAll, UniversalHealthCare, MentalHealthMatters, VaccinesWork, SDG3, HealthyLiving}      \\
            \cmidrule{1-3}
            4                   & Quality Education                       & \texttt{QualityEducation, EducationForAll, InclusiveEducation, LifelongLearning, DigitalLearning, SDG4}             \\
            \cmidrule{1-3}
            5                   & Gender Equality                         & \texttt{GenderEquality, WomenEmpowerment, EndGenderBias, GirlsEducation, SDG5, EqualPay}                            \\
            \cmidrule{1-3}
            6                   & Clean Water and Sanitation              & \texttt{CleanWater, WaterForAll, SanitationMatters, SDG6, SafeWater}                                                \\
            \cmidrule{1-3}
            7                   & Affordable and Clean Energy             & \texttt{CleanEnergy, RenewableEnergy, EnergyAccess, SustainableEnergy, SDG7, GoSolar}                               \\
            \cmidrule{1-3}
            8                   & Decent Work and Economic Growth         & \texttt{DecentWork, InclusiveGrowth, EconomicGrowth, JobCreation, FutureOfWork, SDG8}                               \\
            \cmidrule{1-3}
            9                   & Industry, Innovation and Infrastructure & \texttt{InnovationForAll, SustainableInfrastructure, IndustrialDevelopment, SDG9, TechForGood}                      \\
            \cmidrule{1-3}
            10                  & Reduced Inequalities                    & \texttt{ReducedInequalities, SocialJustice, EqualOpportunities, LeaveNoOneBehind, SDG10}                            \\
            \cmidrule{1-3}
            11                  & Sustainable Cities and Communities      & \texttt{SustainableCities, UrbanDevelopment, SmartCities, ResilientCommunities, SDG11, GreenCities}                 \\
            \cmidrule{1-3}
            12                  & Responsible Consumption and Production  & \texttt{ResponsibleConsumption, SustainableProduction, CircularEconomy, WasteReduction, SDG12, EcoFriendly}         \\
            \cmidrule{1-3}
            13                  & Climate Action                          & \texttt{ClimateAction, ActOnClimate, NetZero, CarbonNeutral, SDG13, ClimateCrisis}                                  \\
            \cmidrule{1-3}
            14                  & Life Below Water                        & \texttt{LifeBelowWater, SaveOurOceans, MarineConservation, PlasticPollution, SDG14, ProtectOurSeas}                 \\
            \cmidrule{1-3}
            15                  & Life on Land                            & \texttt{LifeOnLand, ForestConservation, Biodiversity, SaveNature, SDG15, EcosystemRestoration}                      \\
            \cmidrule{1-3}
            16                  & Peace, Justice and Strong Institutions  & \texttt{PeaceAndJustice, RuleOfLaw, EndCorruption, HumanRights, SDG16, SocialCohesion}                              \\
            \cmidrule{1-3}
            17                  & Partnerships for the Goals              & \texttt{PartnershipsForGoals, GlobalCooperation, TogetherForSDGs, PublicPrivatePartnerships, SDG17, GlobalGoals}    \\
            \bottomrule
        \end{tabular}
    }
\end{table}

\section{GICS Sector Definitions}\label{appendix:gics_sectors}

Provided here are formal definitions for the Global Industry Classification Standard (GICS) sectors as of March 17, 2023.
\begin{description}

    \item [Energy Sector]
          The Energy Sector comprises companies engaged in exploration \& production, refining \& marketing, and storage \& transportation of oil \& gas and coal \& consumable fuels. It also includes companies that offer oil \& gas equipment and services.

    \item [Materials Sector]
          The Materials Sector includes companies that manufacture chemicals, construction materials, forest products, glass, paper and related packaging products, and metals, minerals and mining companies, including producers of steel.

    \item [Industrials Sector]
          The Industrials Sector includes manufacturers and distributors of capital goods such as aerospace \& defense, building products, electrical equipment and machinery and companies that offer construction \& engineering services. It also includes providers of commercial \& professional services including printing, environmental and facilities services, office services \& supplies, security \& alarm services, human resource \& employment services, research \& consulting services. It also includes companies that provide transportation services.

    \item [Consumer Discretionary Sector]
          The Consumer Discretionary Sector encompasses those businesses that tend to be the most sensitive to economic cycles. Its manufacturing segment includes automobiles \& components, household durable goods, leisure products and textiles \& apparel. The services segment includes hotels, restaurants, and other leisure facilities. It also includes distributors and retailers of consumer discretionary products.

    \item [Consumer Staples Sector]
          The Consumer Staples Sector comprises companies whose businesses are less sensitive to economic cycles. It includes manufacturers and distributors of food, beverages and tobacco and producers of non-durable household goods and personal products. It also includes distributors and retailers of consumer staples products including food \& drug retailing companies.

    \item [Health Care Sector]
          The Health Care Sector includes health care providers \& services, companies that manufacture and distribute health care equipment \& supplies, and health care technology companies. It also includes companies involved in the research, development, production and marketing of pharmaceuticals and biotechnology products.

    \item [Financials Sector]
          The Financials Sector contains companies engaged in banking, financial services, consumer finance, capital markets and insurance activities. It also includes Financial Exchanges \& Data and Mortgage REITs.

    \item [Information Technology Sector]
          The Information Technology Sector comprises companies that offer software and information technology services, manufacturers and distributors of technology hardware \& equipment such as communications equipment, cellular phones, computers \& peripherals, electronic equipment and related instruments, and semiconductors and related equipment \& materials.

    \item [Communication Services Sector]
          The Communication Services Sector includes companies that facilitate communication and offer related content and information through various mediums. It includes telecom and media \& entertainment companies including producers of interactive gaming products and companies engaged in content and information creation or distribution through proprietary platforms.

    \item [Utilities Sector]
          The Utilities Sector comprises utility companies such as electric, gas and water utilities. It also includes independent power producers \& energy traders and companies that engage in generation and distribution of electricity using renewable sources.


\end{description}

\section{Prompts for the SDG Classification and Cluster Summarization Tasks}

This section provides the prompts used for the LLM-based SDG annotation process (detailed in \Cref{section:approach_part_one}) and for generating summary descriptions and concepts for the visual understanding task (detailed in \Cref{section:approach_part_two}).

\subsection{LLM prompt for Classifying Tweets into Sustainable Development Goals}\label{appendix:sdg_classification_prompt}
\fbox{%
    \footnotesize
    \ttfamily
    \parbox{\textwidth}{%
        You are an expert in text classification and the United Nations Sustainable Development Goals (SDGs). \\

        Your task is to read a tweet and determine which ONE of the 17 SDGs it most closely aligns with,
        based on its content. If it does not relate to any of the goals, respond with ``None.'' \\

        Here are the 17 SDGs: \\

        1. Goal 1: No Poverty: End poverty in all its forms everywhere. \\
        2. Goal 2: Zero Hunger: End hunger, achieve food security and improved nutrition, and promote sustainable agriculture. \\
        3. Goal 3: Good Health and Well-being: Ensure healthy lives and promote well-being for all at all ages. \\
        4. Goal 4: Quality Education: Ensure inclusive and equitable quality education and promote lifelong learning opportunities for all. \\
        5. Goal 5: Gender Equality: Achieve gender equality and empower all women and girls. \\
        6. Goal 6: Clean Water and Sanitation: Ensure availability and sustainable management of water and sanitation for all. \\
        7. Goal 7: Affordable and Clean Energy: Ensure access to affordable, reliable, sustainable and modern energy for all. \\
        8. Goal 8: Decent Work and Economic Growth: Promote sustained, inclusive and sustainable economic growth, full and productive employment and decent work for all. \\
        9. Goal 9: Industry, Innovation and Infrastructure: Build resilient infrastructure, promote inclusive and sustainable industrialization and foster innovation. \\
        10. Goal 10: Reduced Inequalities: Reduce inequality within and among countries. \\
        11. Goal 11: Sustainable Cities and Communities: Make cities and human settlements inclusive, safe, resilient and sustainable. \\
        12. Goal 12: Responsible Consumption and Production: Ensure sustainable consumption and production patterns. \\
        13. Goal 13: Climate Action: Take urgent action to combat climate change and its impacts. \\
        14. Goal 14: Life Below Water: Conserve and sustainably use the oceans, seas and marine resources for sustainable development. \\
        15. Goal 15: Life on Land: Protect, restore and promote sustainable use of terrestrial ecosystems, sustainably manage forests, combat desertification, and halt and reverse land degradation, and halt biodiversity loss. \\
        16. Goal 16: Peace, Justice and Strong Institutions: Promote peaceful and inclusive societies for sustainable development, provide access to justice for all and build effective, accountable and inclusive institutions at all levels. \\
        17. Goal 17: Partnerships for the Goals: Strengthen the means of implementation and revitalize the Global Partnership for Sustainable Development. \\

        Classification Criteria: \\
        - Identify the main theme of the tweet. \\
        - Match the main theme to the most relevant SDG from the list above. \\
        - If multiple SDGs seem relevant, choose the single best match. \\
        - If the content does not relate to any SDG, respond with ``None.'' \\

        Finally, output only the SDG number (1-17) or ``None''. \\
        Do not provide extra text, formatting, or explanations.
    }%
}

\subsection{VLM Prompt for Generating Visual Summary and Concepts}\label{appendix:visual_summary_prompt}
\fbox{%
    \footnotesize
    \ttfamily
    \parbox{\textwidth}{%
        You are provided with a set of images. Your task is twofold: \\

        1. Generate a single-line, high-level summary of the overall theme of these images. \\
        2. Identify and list each key theme or concept in these images as short phrases in a bullet list. \\

        Format your response as follows: \\
        - On the first line: the single-line summary. \\
        - Followed by bullet points for the key themes. \\

        Provide no additional text or explanation.
    }
}

\section{Spearman Rank Correlations between Proportion of SDG-focussed Communication and ESG Risk}\label{appendix:correlations}

\Cref{tab:appendix_corr1} and \Cref{tab:appendix_corr2} present Spearman rank correlation scores between the proportion of SDG-focused messaging (calculated as the
number of tweets annotated for relevance to a specific SDG divided by a company's total tweet volume) and its ESG risk score.

\begin{table}[h]
    \sisetup{detect-weight,     
        mode=text,         
        table-format=-1.4, 
        add-integer-zero=false,
        table-space-text-post={*} 
    }
    \caption{Spearman Rank Correlation analysis examining the relationship between the proportion of tweets focused on each Sustainable Development Goal (SDG) 1--9---defined as
        the ratio of tweets deemed relevant to a specific SDG to total tweet volume---and overall ESG Risk. Significant correlations ($p < 0.05$) are indicated in \textit{boldface}.}
    \label{tab:appendix_corr1}
    \resizebox{0.95\textwidth}{!}{%
        \begin{tabular}{lSSSSSSSSS}
            \toprule
            Industry               & 1                & 2                & 3                & 4                & 5                & 6                & 7                & 8                & 9                \\ \midrule
            Communication Services & -0.3828          & -0.2860          & -0.4462          & 0.0989           & 0.0022           & -0.1778          & 0.1144           & 0.1297           & 0.4198           \\
            Consumer Discretionary & -0.0565          & 0.1693           & 0.0615           & -0.0067          & -0.0257          & -0.1271          & -0.1362          & 0.0013           & -0.1237          \\
            Consumer Staples       & 0.0019           & 0.1492           & -0.2815          & -0.0408          & -0.0908          & -0.0815          & 0.0062           & -0.0462          & 0.1439           \\
            Energy                 & 0.0258           & 0.3053           & 0.0000           & 0.3965           & \textbf{0.4825}  & 0.3684           & -0.1456          & 0.2772           & -0.4491          \\
            Financials             & \textbf{0.3301}  & 0.2112           & 0.0457           & \textbf{0.2960}  & 0.1470           & 0.0389           & \textbf{-0.2895} & \textbf{0.3037}  & \textbf{-0.3432} \\
            Healthcare             & 0.2166           & -0.1072          & \textbf{0.3579}  & -0.0308          & 0.1160           & \textbf{-0.4191} & -0.0803          & -0.0545          & \textbf{-0.3128} \\
            Industrials            & \textbf{-0.2045} & \textbf{-0.2682} & 0.0304           & -0.0128          & 0.0230           & -0.1188          & 0.0766           & \textbf{-0.2135} & \textbf{0.2747}  \\
            Information Technology & -0.0412          & -0.2045          & \textbf{-0.2839} & \textbf{-0.2723} & \textbf{-0.2846} & \textbf{-0.2699} & -0.0060          & 0.1051           & 0.0541           \\
            Materials              & 0.1655           & 0.2402           & 0.0894           & 0.2419           & 0.1299           & 0.2476           & -0.0918          & \textbf{0.3404}  & -0.2502          \\
            Utilities              & 0.3479           & 0.2569           & 0.2109           & 0.2171           & -0.3785          & 0.0961           & 0.0537           & -0.0824          & 0.0776           \\ \bottomrule
        \end{tabular}
    }
\end{table}

\begin{table}[h]
    \sisetup{detect-weight,     
        mode=text,         
        table-format=-1.4, 
        add-integer-zero=false,
        table-space-text-post={*} 
    }
    \caption{Spearman Rank Correlation analysis examining the relationship between the proportion of tweets focused on each Sustainable Development Goal (SDG) 10--17---defined as
        the ratio of tweets deemed relevant to a specific SDG to total tweet volume---and overall ESG Risk. Significant correlations ($p < 0.05$) are indicated in \textit{boldface}.}
    \label{tab:appendix_corr2}
    \resizebox{0.95\textwidth}{!}{%
        \begin{tabular}{lSSSSSSSSS}
            \toprule
            Industry               & 10      & 11              & 12               & 13               & 14      & 15               & 16      & 17               & All              \\ \midrule
            Communication Services & 0.1077  & 0.2132          & -0.5281          & -0.1648          & -0.4356 & -0.3011          & -0.1956 & -0.1560          & 0.2264           \\
            Consumer Discretionary & 0.0769  & 0.1211          & -0.0383          & -0.0114          & 0.0169  & 0.0692           & -0.0234 & 0.0586           & 0.0525           \\
            Consumer Staples       & 0.0054  & -0.0362         & 0.1223           & -0.1035          & -0.0843 & 0.1110           & -0.1262 & -0.1339          & -0.0500          \\
            Energy                 & 0.0270  & -0.1088         & 0.1842           & \textbf{0.5789}  & -0.3080 & 0.1316           & 0.4404  & 0.2053           & -0.1035          \\
            Financials             & 0.0356  & -0.0357         & -0.1634          & \textbf{-0.2384} & -0.2170 & 0.0238           & 0.1545  & -0.0625          & \textbf{-0.2566} \\
            Healthcare             & -0.0804 & -0.0849         & \textbf{-0.3895} & -0.2232          & -0.1804 & -0.1995          & -0.0080 & 0.0244           & \textbf{0.3735}  \\
            Industrials            & -0.1302 & -0.1323         & \textbf{-0.2947} & -0.0287          & 0.1187  & -0.0133          & -0.0435 & -0.1110          & -0.0032          \\
            Information Technology & -0.1508 & -0.2096         & -0.1418          & \textbf{-0.3190} & -0.2098 & \textbf{-0.2823} & -0.1135 & \textbf{-0.2676} & 0.2057           \\
            Materials              & 0.1469  & 0.2973          & \textbf{-0.3336} & -0.0995          & 0.1658  & 0.1269           & 0.2109  & -0.0944          & -0.0752          \\
            Utilities              & 0.0113  & \textbf{0.4427} & 0.1446           & -0.1391          & -0.2314 & 0.0393           & -0.1412 & -0.1070          & -0.3552          \\ \bottomrule
        \end{tabular}
    }
\end{table}